# HATE SPEECH CLASSIFICATION IN ROMAN URDU: A COMPARATIVE STUDY ON PARAMETER EFFICIENT FINE-TUNING AND PROMPT ENGINEERING

by

Toneema Zubair

MSDS22021

A thesis submitted in partial fulfillment of the requirements for the degree of Master of Science in Data Science

Examination Committee: **Dr. Faisal Kamiran** (Supervisor)
**Dr. Waqas Sultani** (Committee Member)
**Dr. Asim Kareem** (Committee Member)

Department of Computer Science
Information Technology University, Lahore
Pakistan
July 2025

## Acknowledgments

First and foremost, I would like to express my deepest gratitude to **Almighty Allah**, who granted me the strength, patience, determination, and endurance to complete this research successfully.

I am profoundly thankful to my supervisor, **Dr. Faisal Kamiran**, for his continuous guidance, insightful advice, and constant motivation, which have played a vital role in shaping this work and bringing me to where I stand today.

I would also like to extend my heartfelt thanks to **Dr. Hafiz Hassan Saeed**, who has been a mentor throughout this journey. His academic guidance and ethical support have been invaluable at every stage of this research.

My sincere appreciation goes to my friends and colleagues, whose encouragement and motivation kept me focused whenever I lost track.

Last but certainly not least, I owe my deepest thanks to my beloved family, especially my husband, **Zubair Rasheed**, and my three children, for their unwavering support, patience, and love. Without them, this achievement would not have been possible.

# Table of Contents

# List of Figures

## List of Tables

# Abstract

Due to the widespread accessibility of the internet and social media, toxic and hateful content has grown exponentially, causing significant distress and negative societal impacts. Roman Urdu, a low-resource language used in Pakistan and among Urdu-speaking communities worldwide, presents additional challenges because of its informal grammar, inconsistent sentence structures, and multiple variations in word spellings. This research aims to identify the most effective techniques for hate speech classification in such low-resource settings with limited data. To address this, the study investigates and compares the latest approaches, including prompt tuning, parameter-efficient fine-tuning (PEFT) using LoRA, and prompt engineering, under various experimental configurations. To achieve this objective, four experiments were designed. The first experiment involved direct inferencing with LLMs without any fine-tuning, to evaluate how well these models understand Roman Urdu in a zero-shot setting, especially given limited data. The second experiment utilized parameter-efficient fine-tuning (PEFT) with LoRA, which updates only a small subset of parameters, thereby reducing computational cost. The third experiment explored prompt tuning with both mixed and manually crafted prompts, using very small sets of training examples relative to the entire dataset, making it computationally efficient as well. Finally, the fourth experiment applied prompt engineering through zero-shot and few-shot learning, relying solely on carefully designed instruction prompts for classification without further training. The results demonstrate that while prompt engineering can achieve remarkable performance in certain cases, PEFT consistently outperforms all other techniques when considering overall performance and generalization across the entire dataset. This thesis highlights the importance of combining prompt-based approaches with parameter-efficient fine-tuning methods to achieve robust performance in low-resource languages like Roman Urdu.

# Chapter 1

# Introduction

Hate speech is any message that spreads hate, unfairness, or encourages violence against people or groups based on their race, religion, gender, or ethnicity. Its identification is a significant challenge in natural language processing (NLP) as it poses a substantial danger to social cohesion and online safety.

The goal and complexity of the detection of hate speech are different from those of sentiment analysis. Hate speech identification looks for language that is hurtful, abusive, or discriminatory, whereas sentiment analysis categorizes content according to its emotional tone as positive, negative, or neutral. It is possible for a comment to be critical without being hostile. For example:

*"Is mobile ki battery bohat bekaar hai!"*

*English Translation:*

*"The battery of this mobile is terrible!'*

Despite its critical tone, this above-mentioned comment is not hate speech because it critiques a product without naming specific people or groups.

In low-resource languages like Roman Urdu, which frequently lack regular grammar and structure, it is more important to have a better understanding of language, intent, and context in order to accurately distinguish between hate speech detection and sentiment analysis. Conventional sentiment analysis techniques cannot adequately capture the complex and often context-dependent character of hate speech.

## 1.1 Overview

More people are creating informal content due to the widespread use of digital platforms in diverse cultures such as Pakistan. Much of this content is written in Roman Urdu, which is a version of Urdu that uses the Latin alphabet instead of traditional script. Although many people use Roman Urdu on social networks, Natural Language Processing (NLP) systems have considerable difficulties due to their irregular spelling, phonetic variances, and contextual ambiguity.

A major obstacle is the lack of standardized orthography. Depending on how a user interprets phonetics, a word can have several different spellings. For example, in Roman Urdu, the word for "good" in Urdu can be written as acha, achaa, achha, or achaw, all of which convey the same meaning.

Semantic ambiguity, in which a single Roman Urdu word may have several meanings depending on the context, is another significant issue. For example:

Depending on usage, the word *badal* can signify a cloud (transliterated as *baadal*) or change (*badal*).

Automated models might incorrectly categorize certain terms in the absence of contextual signals,

particularly in brief or informal texts. For NLP applications like hate speech identification, where accurate sentiment, tone, and purpose interpretation are essential, Roman Urdu is a challenging language due to these issues. Roman Urdu lacks the structured input and rich linguistic resources that traditional text categorization techniques frequently require.
This thesis investigates contemporary NLP methods that are more appropriate for informal and low-resource languages in order to handle these complications. Specifically, it looks into:

- **Prompt Tuning:** This technique keeps the pretrained language model frozen and attaches a specially designed prompt with a masked token to each input. The model focuses solely on predicting the masked token, significantly reducing training time and computational cost by avoiding full-model fine-tuning.
- **Prompt Engineering:** This approach relies on the user's ability to manually craft well-structured prompts. These prompts may include zero-shot or few-shot examples that guide the model's behavior without updating any of its internal weights.
- **Parameter Efficient Fine-Tuning (LoRA):** This method introduces a small set of trainable parameters into the model while keeping the majority of the base model frozen. By fine-tuning only low-rank adaptation layers, LoRA achieves high performance with reduced training time and resource consumption.

The aim of this thesis is to conduct a comprehensive comparative analysis of these three techniques using PLMs such as BERT (multilingual) and LLMs such as OPENAI GPT models, Llama, Mistral and Deepseek. The results will help to identify the approach that gives the best trade-off between computational power consumption and performance of the models, hence can be deployed in real time scenario especially where there is low resource data and full fine-tuning may not be feasible.

## 1.2 Problem Statement

Research on natural language processing has given hate speech detection a lot of attention over the past decade. With a significant emphasis on high-resource languages, especially English, numerous studies have used both conventional machine learning methods (like logistic regression and SVMs) and deep learning techniques (like CNNs, LSTMs) as well as the more recent transformer-based models. Large, annotated datasets and standardised linguistic structures have made it possible to investigate a variety of methods for detecting hate speech in English, including new developments in large language models.
Research on Roman Urdu, on the other hand, has mostly focused on using conventional machine learning and deep learning techniques, with little investigation into more recent, parameter-efficient NLP approaches. Since Roman Urdu is becoming more and more common on digital platforms and the demand for scalable hate speech identification in low-resource language situations is expanding, this represents a substantial gap.

The exponential rise in social media usage, particularly in South Asian nations, has led to a sharp rise in offensive and hateful content written in informal languages such Roman Urdu. Despite being widely used on many platforms in India and Pakistan, Roman Urdu is still a low-resource language since it lacks publicly accessible annotated corpora, standardised grammar, and consistent spellings. Roman Urdu's informality, semantic ambiguity, and context-dependent phrasing make it even more difficult to identify hate speech since they frequently make it difficult to distinguish between hate speech and general negativity. Therefore, for models to be useful, they need to catch more subtle linguistic patterns and go beyond simple sentiment classification.

While full-model training or fine-tuning are frequently used in traditional procedures, these techniques are computationally costly and need updating all model parameters, which makes them inappropriate for low-resource settings. On the other hand, lightweight solutions are provided by methods such as Parameter Efficient Fine-Tuning (PEFT), Prompt Engineering, and Prompt Tuning. They drastically cut down on training time and hardware requirements by either freezing the base model or fine-tuning a limited group of parameters. They are therefore perfect for classifying hate speech in low-resource languages like Roman Urdu.

These contemporary methods are becoming more and more effective in high-resource environments, but they have not yet been thoroughly investigated for Roman Urdu hate speech identification. The purpose of this study is to close that gap by assessing and contrasting their computational efficiency and efficacy in handling hate speech classification in this underrepresented linguistic domain.

## 1.3 Objectives

This research aims to address the challenges in hate speech detection and classification in Roman Urdu by evaluating the performance of latest, resource efficient NLP techniques. Therefore the primary objectives are:

- To explore the efficiency of Prompt Engineering, Prompt Tuning, and Parameter Efficient Fine-Tuning (PEFT) for binary classification of Roman Urdu comments.
- To implement these techniques on a variety of pre-trained models, including multilingual BERT, OpenAI GPT (via Azure), and LlaMA-based models such as Mistral and Gemma.
- To evaluate each method using standard classification metrics, accuracy, precision, recall, and F1-score, with a focus on handling class imbalance.
- To compare the performance, efficiency, and resource utilization of modern techniques with conventional full fine-tuning methods.
- To identify the most suitable strategy for hate speech detection in low-resource, informal language settings with limited computational capacity.

## 1.4 Limitations and Scope

**Scope:** With the use of contemporary NLP approaches such as Prompt Engineering, Prompt Tuning, and Parameter Efficient Fine-Tuning (PEFT), this study focusses on the binary classification of Roman Urdu comments as either *toxic* or *non-toxic*. LlaMA-based models like Mistral and Gemma, OpenAI GPT (via Azure), and multilingual BERT are among the pre-trained language models used in the experiments. The main goal is to evaluate these methods' classification effectiveness and computational efficiency, especially when it comes to informal and low-resource languages like Roman Urdu.

**Limitations:**

- The study's dataset is naturally unbalanced, with a disproportionately high proportion of non-toxic remarks versus toxic ones. Inverse class frequency weighting was used to correct class imbalance; however, this does not accurately represent how models could function on a dataset that is naturally balanced. Using synthetic balancing or more extensive data gathering, future research should investigate model behaviour in a more balanced environment.

- The dataset size is still limited, and generalizing the findings confidently requires further data collection and annotation. A larger and more diverse dataset would allow for more robust model evaluation and reduce the likelihood of dataset-specific biases.

- The current research is confined to Roman Urdu. For broader validation, similar experiments should be conducted on other low-resource languages to assess whether these techniques behave consistently across different linguistic and cultural contexts. Comparing cross-lingual performance can help uncover language-specific model sensitivities or general patterns.

- The study is restricted to binary classification, i.e., detecting whether a comment is toxic or non-toxic. However, hate speech can appear in many forms, such as religious, gender-based, or racially motivated hate, which are not differentiated in this research. Future extensions could explore multi-class classification, enabling deeper insight into the model's ability to understand context-specific hate.

- The models in this study classify comments at the sentence level only. They are not designed to extract specific hateful phrases or to identify hate speech within longer texts or documents. Future work could incorporate token-level classification or sequence labeling approaches to provide finer-grained detection.

- This research does not explore hybrid techniques that combine prompt-based and parameter-efficient fine-tuning approaches. Investigating such combinations could reveal performance improvements or complementary strengths across methods.

## 1.5 Thesis Outline

I organize the rest of this dissertation as follows.
In Chapter 2, I describe the literature review.
In Chapter 3, I propose my methodology.
In Chapter 4, I present the experimental results.
Finally, in Chapter 5, I conclude my thesis.

# Chapter 2
# Literature Review

## 2.1 Hate Speech Detection using Machine Learning

Traditional *machine learning (ML)* algorithms have been used extensively in early hate speech detection studies because of their ease of use and efficiency in text classification tasks. These models have shown respectable performance in detecting hate speech in a variety of languages, although they mainly rely on manual feature engineering, including n-grams, TF-IDF vectors, and syntactic or lexical characteristics.

Nasir et al. [1] used the HS-RU-20 dataset to propose a two-stage classification pipeline for Roman Urdu that differentiates between neutral, offensive, and hateful content. On both word-level and character-level characteristics, they tested six traditional algorithms: CNN, Random Forest, K-Nearest Neighbours (KNN), logistic regression, multinomial Na¨ıve Bayes, and SVM. Their logistic regression model demonstrated the value of lightweight models for morphologically rich but low-resource languages such as Roman Urdu, achieving the maximum accuracy of 81% in the neutral-hostile categorisation and 87% in the offensive-hate speech split. In more extensive multilingual research, Haider et al. [2] assessed a number of machine learning algorithms on a multi-class hate speech dataset, including Random Forest, SVM, Na¨ıve Bayes, Logistic Regression, AdaBoost, and Gradient Boosting. With an accuracy of 90.26%, Random Forest outperformed all other approaches using TF-IDF-based n-gram features, demonstrating the power of ensemble methods in robust classification. Similarly, Alaoui et al. [3] utilised a text mining pipeline and the Na¨ıve Bayes classifier to classify tweets in English, with up to 93.06% accuracy across two datasets. By applying several embeddings (Word2Vec, Doc2Vec, FastText) and classifiers (SVM, RF, LR, KNN) to a Chadian lingua franca dataset, Sanoussi et al. [4] expanded this investigation to a multilingual context. The maximum accuracy was achieved when FastText embeddings and SVM were combined, with 95.4% for insult categorisation and 93.9% for hate detection. Boishakhi et al. [5] advanced the conversation by putting out a multi-modal hate speech detection method that combines text, audio, and video elements using machine learning. This study highlighted that although machine learning (ML) does very well on structured text, it is frequently insufficient when used alone to identify hate speech conveyed by tone, sarcasm, or visual signals. Velankar et al. made a significant contribution that summarises the wider difficulties of ML-based hate speech detection [6]. Three levels are identified by their hierarchical framework: *data-level* (e.g., dataset bias, label imbalance), *model-level* (e.g., overfitting, lack of contextual comprehension), and *human-level* (e.g., annotation subjectivity, confusing user intent). Their approach underlines the necessity for scalable, context-aware models, laying the framework for exploring deep learning and big language models in upcoming research. Together, these experiments show that machine learning provides a solid basis for detecting hate speech, particularly when paired with well-designed characteristics. They also highlight shortcomings in managing low-resource lan-guages like Roman Urdu, cross-lingual generalisation, and contextual nuances, opening the door for deeper learning and LLM-based strategies covered in later sections.

## 2.2 Hate Speech Detection using Deep Learning Models

In contrast to traditional machine learning, deep learning methods have demonstrated better performance in hate speech detection due to their ability of automatically learning contextual and semantic features from data.

In a thorough analysis of CNN-based methods using data from Twitter, Miran and Yahia [7] shown that CNN and CNN-derived architectures are very useful for detecting hate speech in English. They emphasised issues including limited cross-linguistic generalisation, data sparsity, and language specificity. To enhance the identification of abusive content, Kothuru and Vijayan [8] presented a Bi-LSTM model that incorporates a soft-plus activation function to capture intricate textual patterns. When compared to baseline Autoencoder and Multi-task learning models, their model showed more precision. Dwivedy and Roy [9] presented a deep feature fusion technique that used LSTM layers and transfer learning in a multimodal architecture to merge text and image features. In order to better capture context in social media posts, their investigation highlighted the importance of using multimodal cues. By employing neural networks and transfer learning to detect hate speech in Bengali and Hindi, Phung and Cloos [10] shown that cross-lingual models may be used with little computational overhead, even in low-resource environments. Toxic comment classification in Roman Urdu was pioneered by Saeed et al. [11], who created a sizable annotated dataset (PURUTT) of more than 72,000 comments. They tested a number of word embedding strategies and trained both deep learning and conventional models. They established a standard for Roman Urdu toxicity classification with their ensemble technique, which obtained an F1 score of 86.35%. Bilal et al. [12] used a Bi-LSTM model with an attention layer and unique word2vec embeddings to extend deep learning for Roman Urdu. Their model received an F1-score of 0.885 after being trained on the recently created RU-HSD-30K dataset. They also showed that the model's robustness on cross-domain data was improved by lexical normalisation of Roman Urdu.

## 2.3 Hate Speech Detection using BERT

Recent developments in transformer-based models, especially BERT, have made great progress in the identification of hate speech in a variety of languages.

Barkhodar et al. [13] tackled the HSD-2Lang 2024 challenge by detecting hate speech in Arabic and Turkish tweets. Their method established the efficacy of BERT for morphologically rich languages and produced noteworthy F1 scores by optimising BERT-based models and carrying out ablation research on preprocessing and data balancing. Jahan et al. [14] provided a thorough analysis of data augmentation methods for hate speech detection in a wider context, assessing older methodologies in addition to BERT and GPT-based algorithms. In order to increase classification performance by 0.7% F1 score and drastically reduce label change, they suggested a contextual cosine similarity filtering mechanism based on BERT. A study by Alatawi et al. [15] examined the application of domain-specific word embeddings in a Bi-LSTM model and contrasted it with a model that was based on BERT. On a balanced dataset, their results showed that BERT outperformed conventional deep learning models, achieving the highest F1-score of 96%. The goal of Yun et al. [16] was to identify hate speech and gender prejudice in Korean datasets. They employed hyperparameter tweaking and logits ensemble models based on BERT. In several Kaggle tasks, their system performed admirably, achieving a top F1-score of 0.7711 for gender bias detection. Using Turkish Twitter data, Bayrak et al. [17] utilised the BERT-Base model. Their model was used in a live setting to enable real-time moderation after it was trained on labelled toxic and non-toxic comments, achieving a test accuracy of 92.53%. Lastly, Rajput et al. [18] used the ETHOS dataset to test static BERT embeddings for hate speech identification. In contrast to more conventional embeddings like FastText and GloVe, their research demonstrated that neural networks containing BERT embeddings performed better and had more specificity. All of these investigations support the efficacy of BERT and its variations in identifying hate speech, particularly in languages with complicated morphology or limited resources.

## 2.4 Hate Speech Detection using Prompt Engineering and LLMs

Large Language Models (LLMs) are being used for a variety of NLP applications, such as hate speech detection, thanks to recent developments in this field. Xu et al. [19] suggested employing QLoRA to fine-tune the LlaMA2-7B model for news topic categorisation, which resulted in higher accuracy than the Roberta and DeBERTa variations. The potential of LLMs when well optimised in data-rich classification situations was proved by this fine-tuning technique. Nadeau et al. [20] assessed several LLMs, such as the LlaMA2, Mistral, Gemma, and GPT models, on the basis of factuality, toxicity, bias, and hallucination in another benchmark research. The trade-offs in LLM safety design were highlighted by the fact that Mistral demonstrated more consistent performance in multi-turn talks with less hallucinations than LlaMA2, which demonstrated great outcomes in toxicity management but a higher susceptibility for hallucinations. Additionally, prompt engineering has been extensively investigated as an effective substitute for complete model fine-tuning. Domain-Enhanced Prompt Learning (DePL) was introduced by Zhang et al. [21] for the identification of implicit hate speech in Chinese. Their approach achieves state-of-the-art outcomes in both few-shot and full-scale tasks by combining domain fusion with fast learning to solve the issues of data scarcity and subtle, implicit hate speech. Using zero-shot multimodal hate speech detection with simplified prompts, Yamagishi [22] showed that these minimum prompts could outperform complicated ones, increasing efficiency and reducing resource usage in unsupervised environments. Prompt-GAN, a radicalised neural network, was suggested by Govers et al. [23]. It generates synthetic extremist data by adversarial prompt tuning. In addition to providing memory and runtime advantages over conventional fine-tuning, the model increases F1-score by up to 10.1%. For hate speech recognition, Han and Tang [24] investigated in-context learning with GPT-3. Their findings demonstrated how crucial prompt quality is for attaining high performance in few-shot circumstances, both in terms of task description and input-label instances. These studies demonstrate a distinct trend towards scalable approaches for hate speech detection, particularly in multilingual and low-resource settings, that rely on effective, prompt-based procedures and LLM fine-tuning techniques.

## 2.5 Most Relevant and Related Studies

The CRU dataset for cybercrime detection in Roman Urdu, organised in accordance with Pakistan's PECA law, was presented by Ullah et al. [25]. Prefix prompts work better in Roman Urdu than cloze-style prompts, demonstrating the usefulness of prompt engineering in low-resource categorisation tasks, according to their tests with multilingual PLMs and prompt engineering. In a further investigation, Ullah et al. [26] examined five Urdu and Roman Urdu datasets and contrasted prompt-based and conventional fine-tuning techniques. In areas with minimal labelled data, prompt-based approaches showed their strength with an accuracy gain of up to 13%. Liu et al. [27] offered a thorough analysis of prompt-based learning in natural language processing. The methodological decisions made in this thesis are supported by their framework, which describes different prompt formulations and tuning techniques that provide efficient few-shot and zero-shot learning using pre-trained models. Saeed et al. [**?**] developed the first large-scale labeled corpus for toxic comment classification in Roman Urdu. Their ensemble modeling approach achieved a benchmark F1-score of 86.35%, providing a valuable foundation for toxic and hate speech research in this under-resourced language.

# Chapter 3

# Methodology

*This chapter describes the thorough process used to categorise hate speech in Roman Urdu. The pipeline as a whole has several steps, beginning with the exploration of the dataset and the management of class imbalance. Next, appropriate evaluation metrics are chosen to gauge the success of the model. The approach then explores three main avenues: (1) prompt tuning with BERT-based models, (2) prompt engineering with OpenAI's GPT-3.5 through Azure services, and (3) parameter-efficient fine-tuning (PEFT) with large language models (LLMs) like LlaMA, Mistral, and DeepSeek using the LoRA technique. Every experimental setting is meticulously created with particular batch processing, quantisation, and training techniques, as well as suitable frameworks and customised prompts. Roman Urdu is used as a case study to assess and contrast contemporary prompting and fine-tuning methods for low-resource languages.*

## 3.1 Dataset Description

The dataset used in this research is the PURUTT (Parallel Urdu and Roman Urdu corpus for Toxic Comments and Transliteration) dataset, developed by Saeed et al. [11]. It is a specialized hate speech corpus designed for toxic comment classification tasks. This dataset is not publicly available and was obtained exclusively for research purposes through a formal request.

PURUTT is considered one of the most comprehensive corpora for Urdu and Roman Urdu hate speech, containing a large number of preprocessed and annotated comments. It has proven to be a valuable resource for classification tasks involving Roman Urdu, a low-resource language that lacks standardized linguistic rules, but is widely used throughout South Asia, particularly in Pakistan and India, on social media and other digital platforms.

The dataset consists of a total of 72,771 comments, categorized into two classes: *toxic* and *non-toxic*. The comments were collected from social media platforms and are available in both Roman Urdu and native Urdu script. For the purpose of this research, only the Roman Urdu comments were utilized.

**Table 3.1:** Summary of the PURUTT Dataset

| Class | Number of Samples | Class Weight |
|---|---|---|
| Toxic | 13,097 | 0.8202 |
| Non-Toxic | 59,674 | 0.1792 |
| **Total** | **72,771** | 1 |

The dataset exhibits a significant class imbalance, with 13,097 comments labeled as toxic and 59,674 as non-toxic. To address this imbalance during model training, inverse class frequency weighting was applied. The calculated class weights are as follows:

- Toxic class weight: 0.8202
- Non-toxic class weight: 0.1792

This weighting strategy was used to reduce the bias introduced by the majority class and to strengthen the contribution of the minority class in the learning process.
In addition to the imbalanced distribution of the dataset, Roman Urdu poses unique linguistic challenges, including inconsistent spellings, phonetic variations, and lexical ambiguity. These characteristics demand a comprehensive understanding of contextual meaning, making the dataset an ideal test case for evaluating the effectiveness of the proposed techniques. Furthermore, the findings derived from this study are directly applicable to real-world scenarios involving informal and low-resource language processing.

## 3.2 Imbalanced Dataset and Its Challenges

An essential part of training successful machine learning models is managing imbalanced datasets, particularly in binary classification situations when one class is noticeably under-represented. There is a notable class imbalance in our dataset, with a considerably greater proportion of non-toxic remarks than toxic ones. This imbalance may result in biased learning and poor generalisation on minority class predictions. There are a number of algorithm-level strategies to deal with this imbalance. These include threshold-moving approaches, one-class learning, cost-sensitive learning, random undersampling of the majority class, and synthetic oversampling techniques like SMOTE [28]. Every strategy has unique benefits and drawbacks. For example, in informal text domains like Roman Urdu, where language inconsistencies make it challenging to generate high-quality synthetic data, oversampling can lead to overfitting. Conversely, undersampling eliminates potentially useful information from the majority class, which could restrict the model's capacity to be applied broadly. Cost-sensitive learning strategy was used in this study. In particular, class weighting was used to penalise the minority class (toxic comments) for misclassification more severely than the majority class (non-toxic comments). This method encourages the model to focus more on under-represented samples while preserving the entire dataset. When the dataset size is enormous but unbalanced, as with PURUTT, it works particularly well. The effectiveness of cost-sensitive learning in situations where changing the class distribution is undesirable or impractical [28] is described by Sotiris Kotsiantis et al., which supports this decision. Class weights are added to the loss function during training, which teaches our model to modify its decision boundary to better handle minority class examples without compromising performance.

### 3.2.1 Class Weight Calculation

The class weights were computed using the standard formula:

$$\text{Weight}_i = \frac{N}{C \times N_i} \quad \text{(Equation 3.1)}$$

Where:

- $N$ is the total number of samples in the dataset
- $C$ is the total number of classes (in this case, 2)
- $N_i$ is the number of samples in class $i$

Substituting values:

$$\text{Weight}_{\text{toxic}} = \frac{72771}{2 \times 13097} \approx 2.78$$

and

$$\text{Weight}_{\text{non-toxic}} = \frac{72771}{2 \times 59674} \approx 0.61$$

These values were then normalized:

$$\text{Normalized Weight}_{\text{toxic}} = \frac{2.78}{2.78 + 0.61} \approx 0.8202$$

and

$$\text{Normalized Weight}_{\text{non-toxic}} = \frac{0.61}{2.78 + 0.61} \approx 0.1798$$

These normalized weights were incorporated during training to ensure that the model pays appropriate attention to the minority class.

### 3.2.2 Evaluation Metric: F1-score

The use of the F1-score is emphasized as the main evaluation metric in addition to weighted loss. Accuracy alone may be deceptive in imbalanced classification problems, where the F1-score (the harmonic mean of precision and recall) is especially helpful. As a result, the evaluation of model performance in both classes was more evenly distributed [29, 30, 31, 32, 33]

## 3.3 Investigations

This work is organised along three main lines of inquiry in order to methodically assess the efficacy of different contemporary NLP algorithms for hate speech identification in Roman Urdu:

- Comparison of Parameter-Efficient Fine-Tuning with Multilingual BERT, Prompt Tuning, and Direct Inferencing: Using the multilingual BERT (uncased) model, this study investigates the performance differences between parameter-efficient fine-tuning (PEFT), lightweight prompt tuning, and zero-shot inferencing. The goal is to comprehend how classification performance on Roman Urdu data is impacted by varying degrees of model adaption.

- Analysing Large Language Models (LLMs) using Parameter-Efficient Fine-Tuning and Direct Inferencing: This study evaluates the fine-tuning effectiveness and direct inferencing capabilities of a number of open-source LLMs on Roman Urdu text, including LlaMA3.2, Mistral, Gemma, DeepSeek, and Falcon. These tests aid in determining which models are most suited for low-resource downstream categorisation jobs.

- OpenAI Models: Zero-Shot vs. Few-Shot Prompt Engineering: This method uses quick engineering to assess the performance of commercial LLMs retrieved through OpenAI's API. Insights into the prompt sensitivity of generative models on informal and code-mixed text are provided by comparing zero-shot and few-shot prompting strategies for the classification of toxic comments in Roman Urdu.

The goal of these studies is to offer a thorough assessment of transformer-based techniques across various adaption paradigms, which will be the basis for choosing effective and precise strategies suited to low-resource textual material, such as Roman Urdu.

### 3.4 Prompt Tuning with Manual and Soft Templates

The prompt tuning process for Roman Urdu toxicity classification is described in this section. I have presented the OpenPrompt framework, explained the Roman Urdu verbalizer that translates the `[MASK]` prediction to class labels, and defines two prompt families: *manual* templates (prefix, cloze) and *mixed* templates (manual + learnable soft tokens). I also described how these templates are implemented in the experiments and how they correspond to BERT's MLM goal.

#### 3.4.1 Framework and Setup

The **OpenPrompt** [34] framework was used for prompt tuning; it supports a variety of prompt styles, such as manual, soft, and mixed templates, and interfaces smoothly with Hugging Face's ecosystem of PLMs. Additionally, OpenPrompt offers a variety of customisable verbalisers, including automated, knowledgeable, and manual ones, enabling versatile label-word mapping experiments.
For prompt tuning experiments, `BERT-base-multilingual-uncased` was the pre-trained language model. This model is frequently utilised in multilingual NLP jobs and supports Roman Urdu.

#### 3.4.2 Prompt Styles Used

This study explores two broad categories of prompt tuning:

- **Manual Prompt Tuning:** Two types of manually constructed prompts were tested:
  - *Prefix Prompt:* e.g., `Ye hai [MASK]!`
  - *Cloze Prompt:* e.g., `Ye [MASK] hai.`

  These prompt formats make explicit use of the `[MASK]` token, which instructs the model to create or categorise the masked position according to the context. Human-designed manual prompts provide control over task framing and syntactic structure, which is essential for informal and under-resourced languages like Roman Urdu. This method was motivated by Ullah et al. [26], who demonstrated that prompt-based fine-tuning performed better than traditional fine-tuning in tasks involving the classification of texts in Urdu and Roman Urdu. Our implementation expanded the evaluation while adhering to their manual strategy.

- **Mixed Prompt Tuning (New Contribution):** This study's use of mixed stimuli is an innovative approach. In this method, trainable soft tokens are added to human-written templates that contain a `[MASK]` token. The model is able to learn optimised task-specific instructions without changing the base PLM since these tokens are initialised as learnable vectors and modified via backpropagation during training.

  An example of the mixed prompt format is:

  ```
  {"placeholder":"text_a","soft":"Yeh","soft":"comment hai","mask"}
  ```

  Soft tokens enable the prompt to encode latent task-relevant signals, and the `[MASK]` token supports the transformer's masked language modelling (MLM) goal. This hybrid approach improves task generalisability in low-resource contexts, such as Roman Urdu toxicity categorisation, by utilising both human intuition and model optimisation.

### 3.4.3 Manual Verbalizer

To translate the predicted word at the mask position into a final class label (toxic or non-toxic), a manual verbalizer was used. This dictionary associates class labels with semantically significant words (such as ”ghussa” and ”muhabbat”). The socio-linguistic features of Roman Urdu discourse were taken into consideration when choosing these label words.

```
{ "0": ["muhabbat", "pyaar", "acha", "dua", "wah"],
"1": ["khilaf", "nafrat", "ghussa", "lanat", "ganda", "beh***d",
"gaali", "bura"] }
```

### 3.4.4 Expanded Description: Manual Prompts

Using pre-defined, preset textual templates with a `[MASK]` token positioned strategically to direct the pre-trained language model (PLM) in generating label predictions is known as manual prompt tuning. With this method, a classification challenge is efficiently reformulated as a masked language modelling (MLM) problem.

**Prefix Prompt Format:** `Ye hai [MASK]!`
The sentence ends with the `[MASK]` token. A term from the verbaliser list, like "acha" or "'bura," which semantically represent the class labels, is used by the model to fill this space. These words are chosen from a pre-made verbaliser that assigns each class to semantically comparable terms. This structure makes use of contextual cues preceding the mask and fits in nicely with PLMs' sentence completion logic. This could be because the linear sentence completion form more closely matches the pretraining goal of BERT.

**Cloze Prompt Format:** `Ye [MASK] hai.`
In order to mimic natural sentence structures, the `[MASK]` token is embedded within the sentence. Richer contextual cues from both sides of the mask are provided by this method, which could increase prediction accuracy.

**Significance of the [MASK] Token** In models such as BERT, the `[MASK]` token is essential to the MLM(Mask Language Modelling) goal. The model is encouraged to produce a label word that most accurately captures the sentiment or toxicity of the input comment when the mask is embedded within a skilfully written prompt. For low-resource languages with less labelled data, such as Roman Urdu, this is very effective.
Figure 3.1 illustrates the working mechanism of both cloze and prefix prompt designs and their interaction with the verbaliser and MLM head.

### 3.4.5 Explanation: Mixed Prompts (Manual + Soft Tokens)

The advantages of both manual and soft prompts are combined in the hybrid technique known as mixed prompt tuning. This method incorporates learnable soft tokens into a manual prompt structure, as opposed to depending only on learnt embeddings or fixed textual templates.

- Uses a `[MASK]` token to maintain human-readable syntax.
- To collect task-relevant aspects, trainable soft tokens are inserted either before or after the text. Because the underlying model is frozen, it facilitates parameter-efficient training.
- The combination of trainable context embedding and quick interpretability enhances generalisation.

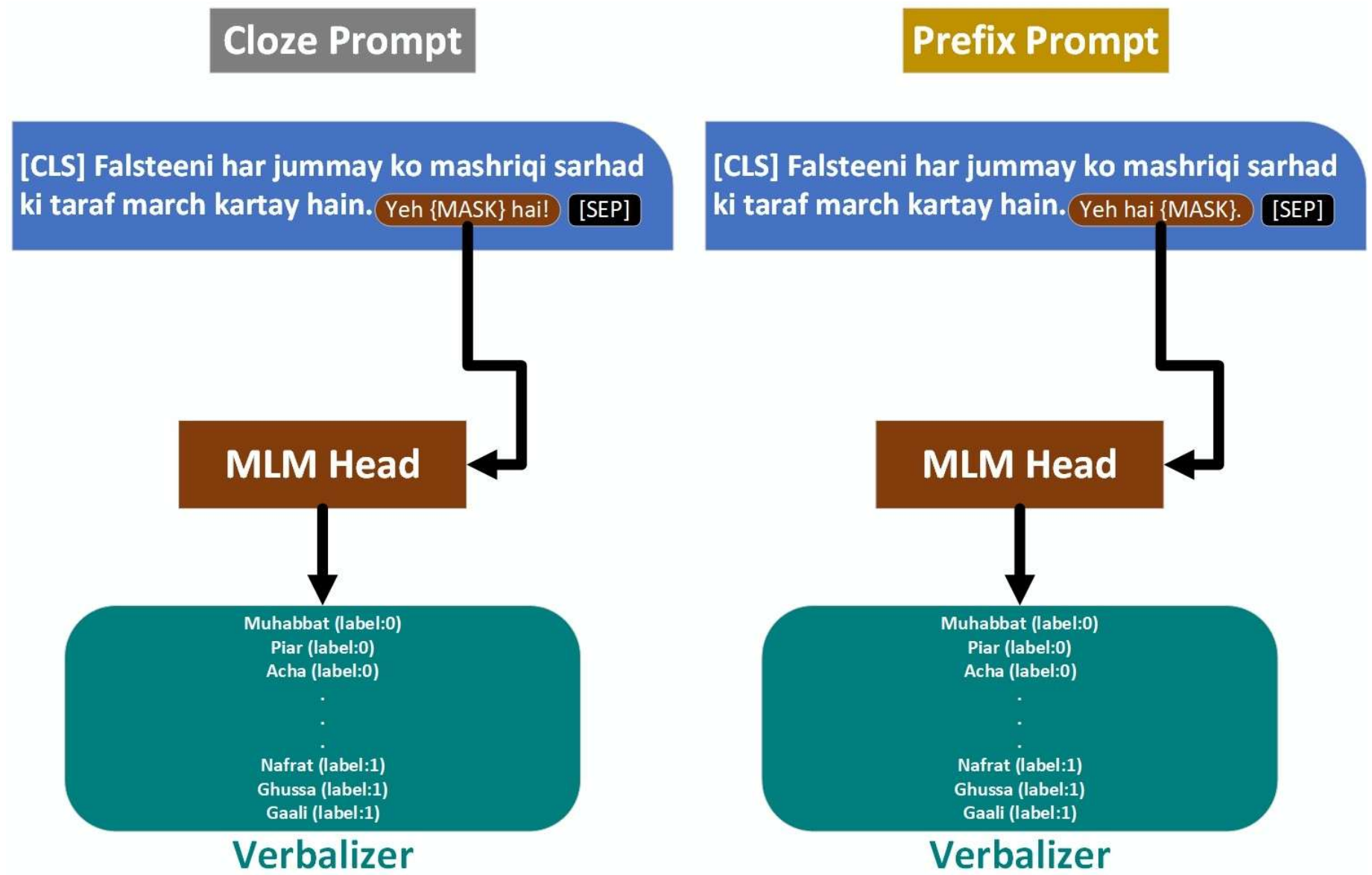

**Figure 3.1:** Visualization of manual prompts: cloze and prefix style, showing how the `[MASK]` token is placed and how the verbaliser interprets the output.

- Few-shot adaptation without overfitting on limited Roman Urdu data is made possible by this item.

**Implementation:**

- OpenPrompt's mixed template support was used to implement it.
- Soft tokens (such as 10–20) are learnt by gradient descent and initialised at random.
- Metrics like accuracy and F1-score were used for evaluation.

Figure 3.2 shows the architecture of the mixed prompt approach, highlighting the incorporation of soft tokens along with manual textual cues.

### 3.4.6 Comparison with Previous Work

Ullah et al. [26] did not look at soft or mixed prompts, but they did concentrate on manual prompt tuning in Urdu and Roman Urdu using both prefix and cloze forms. In addition to replicating their results on manual prompts, our work extends the concept by using mixed prompt tuning, building on their work. A more thorough examination of prompt-based learning in languages with limited resources is made possible by this dual method.

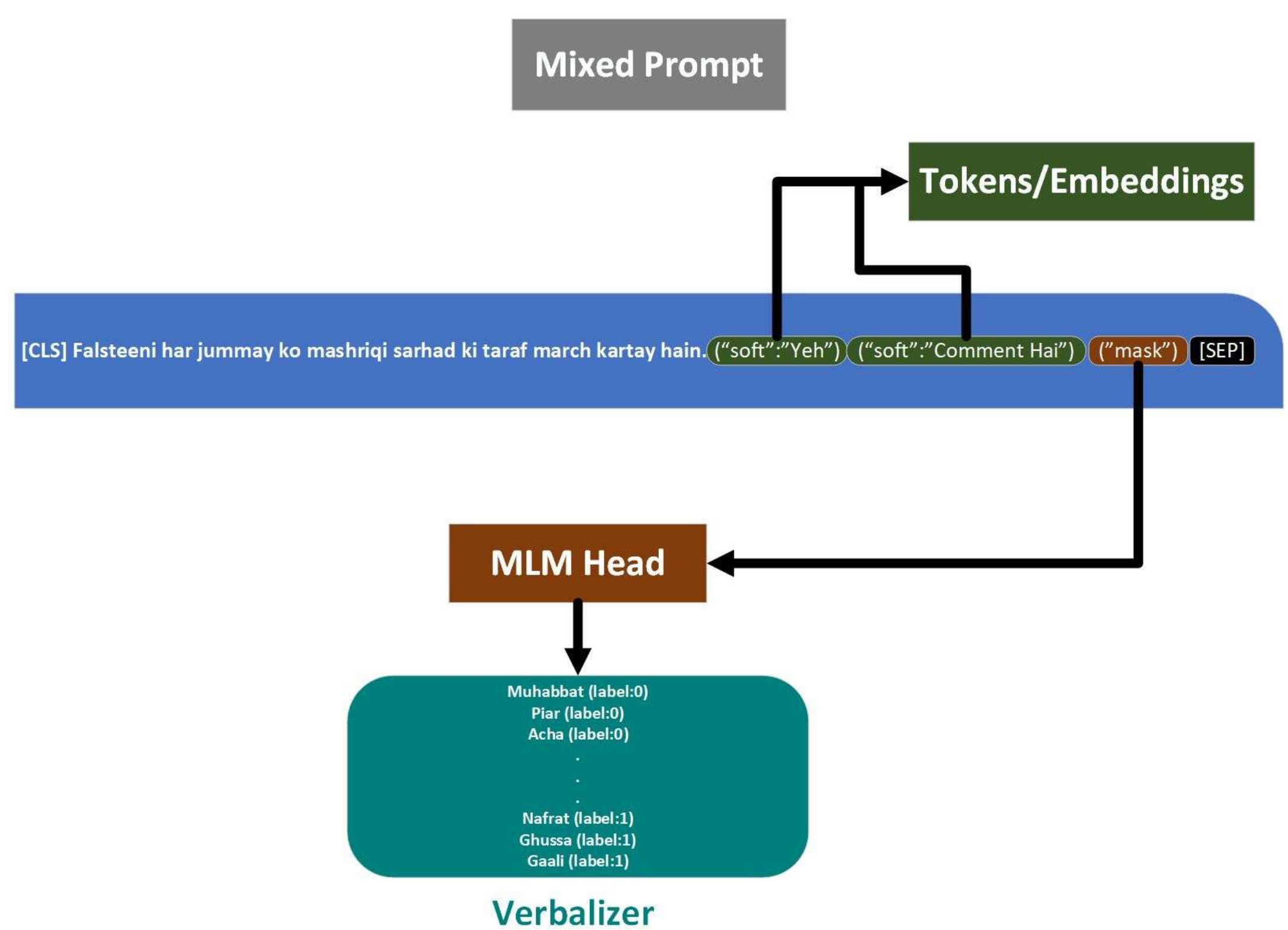

**Figure 3.2:** Mixed prompt structure incorporating both manual templates and soft learnable tokens, demonstrating its connection to the MLM head and verbaliser.

### 3.4.7 Summary of Prompt Styles

Table 3.2 provides a concise side-by-side comparison of the prompting techniques used in this work: manual (prefix / close) and mixed (manual templates augmented with learnable soft tokens). It highlights examples, adaptability, interpretability, cost, and strengths. Use this table as a quick reference to select an appropriate prompt strategy in low-resource Roman Urdu settings.

**Table 3.2:** Comparison of Manual and Mixed Prompt Tuning Strategies

| Aspect | Manual Prompt Tuning | Mixed Prompt Tuning |
|---|---|---|
| Prompt Type | Human-written templates (prefix/cloze) | Manual templates + learnable soft tokens |
| Examples | `Ye hai [MASK]!`, `Ye [MASK] hai.` | `{"soft":"yeh"}, {"soft":"comment hai"}, {"mask"}` |
| Adaptability | Static; task-specific | Dynamic; soft tokens learned |
| Interpretability | High | Moderate (partially interpretable) |
| Cost | Low | Slightly higher |
| Strength | Transparent low-resource use | Flexible, few-shot tuning |

**Algorithm 1** Prompt Tuning Using OpenPrompt Framework

**Input:** $D$: Roman Urdu dataset with comments and labels
**Input:** $PLM$: Pre-trained language model (e.g., BERT-base-multilingual-uncased)
**Input:** $T$: Prompt template style (prefix, cloze, or mixed)
**Output:** Trained prompt-based classifier

1: Split $D$ into $D_{\text{train+dev}}$ (80%) and $D_{\text{test}}$ (20%)
2: Define verbalizer with label words mapped to class 0 and 1
3: **for** sample size $n$ in {32, 64, 128} **do**
4: **for** repetition $r = 1$ to 2 **do**
5: Randomly sample $n$ examples from $D_{\text{train+dev}}$ as $D^{(r)}_{\text{train}}$
6: Use the remaining as $D^{(r)}_{\text{dev}}$
7: Initialize OpenPrompt pipeline with model $PLM$
8: **if** $T$ = prefix **then**
9: Template ← "Ye hai [MASK]!"
10: **else if** $T$ = cloze **then**
11: Template ← "Ye [MASK] hai."
12: **else if** $T$ = mixed **then**
13: Template ← `{"placeholder": "text_a", "soft":"Yeh", "soft":"comment hai", "mask"}`
14: **end if**
15: Create PromptDataLoader for $D^{(r)}_{\text{train}}$
16: Initialize loss and optimizer (for soft tokens if $T$ = mixed)
17: **for** epoch = 1 to 3 **do**
18: **for** each batch in $D^{(r)}_{\text{train}}$ **do**
19: Compute loss and update parameters
20: **end for**
21: **end for**
22: Evaluate on $D^{(r)}_{\text{dev}}$ and store results
23: **end for**
24: Select best model among both repetitions
25: Evaluate best model on $D_{\text{test}}$
26: **end for**

## 3.5 Prompt Engineering using OpenAI GPT-3.5

In this part, I describe the prompt engineering process for the classification of Roman Urdu toxicity using Microsoft Azure OpenAI `GPT-3.5`. I describe the system/user/assistant chat-style message architecture, the two prompting techniques (zero-shot and few-shot), the runtime stack and API setup, and a batching workflow that processes the entire corpus in chunks of 100 comments (728 repetitions per batch). In addition, I outline the evaluation methodology (accuracy, precision, recall, F1), mention practical limitations (such as token restrictions), and give the precise system command. The batch-level results are averaged over all iterations to determine the final results.

### 3.5.1 Libraries and Framework

Microsoft Azure's OpenAI service was used for prompt engineering tests, notably utilizing the `GPT-3.5` model that was launched via Azure endpoint APIs. The required libraries were `openai` and `tiktoken`, and data processing libraries like `pandas`, `sklearn`, and `tqdm` were also included. A Python notebook was used to incorporate the OpenAI client after it was authenticated using API credentials.

### 3.5.2 Prompt Engineering Techniques

Both **zero-shot** and **few-shot** prompting strategies were investigated:

- **Zero-shot:** Without any in-context examples, the model was given a system prompt to categorize comments.
- **Few-shot:** Five instances from each class, toxic and nontoxic, were provided in each prompt and appended before the classification question to improve the model's generalization.

### 3.5.3 Prompt Structure and Formatting

The prompt's structured format consisted of a list of dictionaries, each of which listed a role and the content of its message:

- **System role:** Gave the model instructions tailored to the task.
- **User role:** Carried Roman Urdu comment text.
- **Assistant role:** Contained the label response (0 or 1).

A typical prompt consisted of:

1. A system message explaining the task.
2. For few-shot, a list of user-assistant pairs (input comment and its label).
3. The final comment to classify.

### 3.5.4 Prompt Flowchart

Figure 3.3 illustrates the designed flow of constructing prompts using system instructions, user-assistant examples, and final user query messages, which together guide the model effectively during classification.

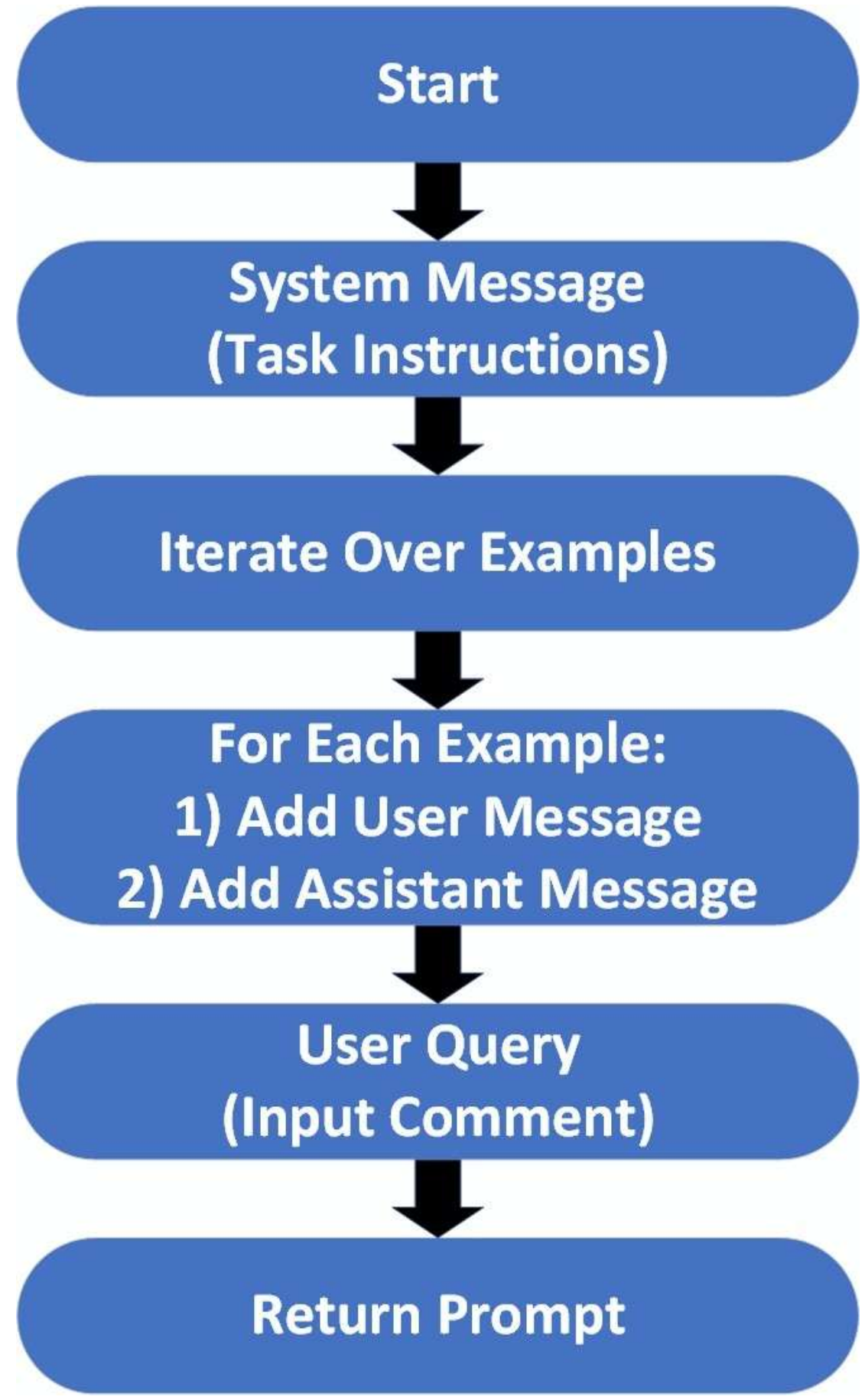

**Figure 3.3:** Flowchart describing prompt construction and message sequence for classification tasks.

### 3.5.5 Batch Processing

In order to evaluate the entire dataset efficiently, prompt evaluation was completed in batches. In particular:

- **Batch Size:** 100 comments per batch.
- **Total Iterations:** 728 iterations were performed for zero-shot and then repeated for few-shot configuration, resulting in 1456 total iterations.
- Responses were processed and saved with expected labels after comments were streamed in JSON format over the OpenAI API.

### 3.5.6 System Prompt Template

**"You are a hate speech detection assistant. You will be provided with a comment. Each comment will be enclosed by triple backticks. You need to read the comment and classify it into one of the below two categories:**
**1 - for toxic comments (containing hate speech, offensive language, or harmful content).**
**0 - for non-toxic comments (free of hate speech and offensive language).**
**Do NOT use any other categories like Neutral or Mixed. If both types of content exist, choose the dominant tone. This is your FINAL CLASSIFICATION. Only output 0 or 1. No explanations."**

### 3.5.7 Implementation Notes

- Prompts were token-limited to `max_tokens = 255` due to GPT-3.5 input constraints.
- Results were parsed using `OpenAI`'s response API and evaluated using accuracy, precision, and recall.
- For few-shot, examples were randomly sampled and refreshed for each batch.

**Algorithm 2** Prompt Engineering using OpenAI for Roman Urdu Hate Speech Detection

**Input:** $D$: Roman Urdu dataset with comments and labels

**Input:** $M$: OpenAI GPT-3.5 model

**Output:** Predicted labels and evaluation metrics

1: Define system message with strict instructions to output only 0 or 1

2: Split dataset $D$ into batches of 100

3: **for** each batch $B_i$ in $D$ **do**

4: Initialize prompt with system message

5: **if** few-shot **then**

6: Select 5 examples per class from $B_i$ and add to prompt

7: **end if**

8: **for** each comment $c$ in $B_i$ **do**

9: Add user message with $c$

10: Model generates predicted label (0 or 1)

11: Save predicted label

12: **end for**

13: Compute batch-level metrics: accuracy, precision, recall, F1-score

14: **end for**

15: Compute final overall metrics as the average of all batch metrics

16: Return labeled dataset and overall metrics

### 3.6 Parameter-Efficient Fine-Tuning (PEFT) using LoRA

This section details a parameter-efficient fine-tuning (PEFT) setup using LoRA for Roman Urdu toxicity classification. The workflow covers (i) backbone models accessed via Hugging Face, (ii) the compute environment (NVIDIA A100 on Google Colab), (iii) a 60/20/20 train–validation–test split, and (iv) LoRA adapter configuration (low-rank $r$, $\alpha$, dropout, bias). The training stack includes the tokenizer, data collator, a class-weighted loss to handle imbalance, and metric logging for accuracy, precision, recall, and F1. Key hyperparameters (learning rate, batch size, epochs, weight decay) are specified, and a schematic contrasts full fine-tuning with LoRA to illustrate how low-rank updates reduce trainable parameters while preserving performance.

#### 3.6.1 LLMs and Hugging Face

The Low-Rank Adaptation (LoRA) technique, which allows fine-tuning large pre-trained language models with minimal trainable parameters, was selected for parameter-efficient fine-tuning experiments. As a result, memory consumption and computational expenses are greatly decreased. Every model was accessible through the Hugging Face API, which facilitates seamless integration with contemporary transformer designs.
This study used LoRA-based PEFT on six distinct pre-trained models:

- Llama 3.2 (8B)
- Gemma2B
- Mistral 7B
- Falcon 7B
- DeepSeek R1-Qwen 7B
- BERT multilingual cased

These models were chosen to investigate the performance of both Pre-trained Language Models (PLMs) and advanced large language models (LLMs).

#### 3.6.2 Hardware and Data Split

Due to limited availability of personal GPU resources, the GPU used for this study was the NVIDIA A100, hosted on Google Colab. This high-performance GPU enabled efficient fine-tuning of large-scale models while ensuring computational feasibility within a cloud environment.
The dataset was split into three subsets:

- 60% for training

- 20% for validation
- 20% for testing

This split strategy guarantees a fair approach to hyperparameter tuning, model fitting, and unbiased evaluation.

### 3.6.3 Quantization and LoRA Adapter Configuration

Specific hyperparameters were set for LoRA to achieve efficient adaptation:

- **Low-rank dimension (*r*):** 16
- **LoRA alpha (scaling factor):** 8
- **LoRA dropout probability:** 0.05
- **Bias:** None
- **Task type:** Sequence classification (`Seq_cls`)

These configurations made it possible to use low-rank updates without having to retrain the entire model weights, allowing the learning of task-specific modifications.

### 3.6.4 Trainer Components and Configuration

The customized trainer incorporated:

- Pre-trained backbone model and tokenizer
- Train and evaluation datasets
- Data collator for dynamic batching
- Explicit metric computation function (`compute metrics`) to automatically log accuracy, precision, recall, and F1-score during training and evaluation

A customized loss function was employed to address the imbalanced nature of the dataset. The weights incorporated in this function were derived directly from the class distribution, ensuring that the model penalizes misclassifications of the minority class more heavily. This approach helps mitigate bias toward the majority class and improves overall fairness in predictions.

### 3.6.5 Training Hyperparameters

The following hyperparameters guided the training procedure:

- Learning rate: $1 \times 10^{-4}$
- Batch size: 8 (train and eval)
- Epochs: 2
- Weight decay: 0.01
- Evaluation and save strategy: per epoch
- Load best model at end: Enabled

### 3.6.6 Summary of PEFT with LoRA Setup

This method allowed for the effective adaption of large-scale transformer models while just changing lightweight LoRA parameters and maintaining the base model frozen. When complete fine-tuning is not feasible, such as in low-resource jobs as Roman Urdu toxin detection, the method is very beneficial.

### 3.6.7 Illustrative Workflow

Figure 3.4 illustrates the comparative workflow between full fine-tuning and LoRA-based parameter-efficient fine-tuning. It highlights how LoRA approximates full rank updates using low-rank decompositions, allowing efficient adaptation while keeping most pretrained weights frozen.

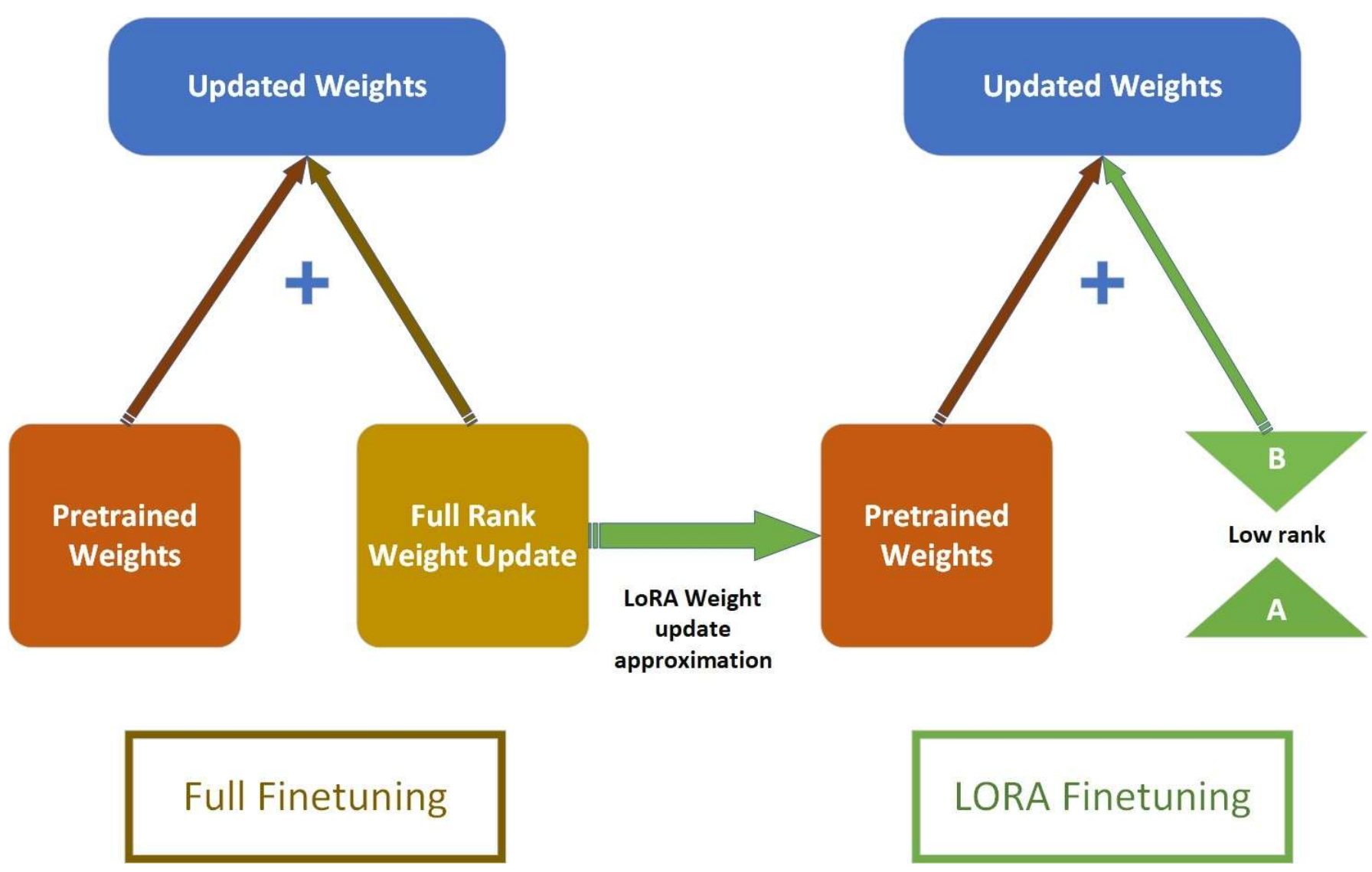

**Figure 3.4:** Comparative workflow of full fine-tuning vs. LoRA fine-tuning. LoRA replaces full-rank weight updates with low-rank approximations (matrices A and B), significantly reducing the number of trainable parameters.

**Algorithm 3** PEFT with LoRA for Roman Urdu Toxicity Classification

**Input:** Roman Urdu dataset $D$, pre-trained model $M$

**Output:** Fine-tuned LoRA model for toxicity classification

1: Split $D$ into $D_{train}$ (60%), $D_{val}$ (20%), and $D_{test}$ (20%)

2: Compute class weights from $D_{train}$ to handle imbalance

3: Tokenize $D_{train}$, $D_{val}$, and $D_{test}$ using $M$'s tokenizer

4: Configure LoRA parameters: $r = 16$, alpha = 8, dropout = 0.05, bias = None, task = Seq_cls

5: Freeze $M$ base parameters and attach LoRA adapters to transformer layers

6: Setup a customized trainer with:

- Weighted loss function (uses computed class weights)
- Tokenizer, data collator, metric computation (Accuracy, Precision, Recall, F1)

7: Define training hyperparameters: LR $= 1 \times 10^{-4}$, batch size = 8, epochs = 2, weight decay = 0.01

8: **for** each epoch **do**

9: Train LoRA parameters on $D_{train}$

10: Evaluate on $D_{val}$, save best checkpoint if improved

11: **end for**

12: Evaluate final model on $D_{test}$ for generalization

13: Report final metrics: Accuracy, Precision, Recall, F1-score

# Chapter 4

# Experimentation and Results

*Four experiments were conducted using different strategies to assess the performance of large language models (LLMs) for hate speech classification in Roman Urdu under various conditions. This section provides a detailed discussion of each experiment, highlighting their methodologies and corresponding results.*

## 4.1 Baseline LLM Performance Evaluation

This experiment establishes a baseline by running direct (zero-shot) inference with quantized Hugging Face LLMs on the held-out test set. Models were evaluated without any fine-tuning, using batched, gradient-free inference with a maximum sequence length of 512 tokens. The goal is to provide a reference point for later adapted methods (prompt tuning and PEFT).

### 4.1.1 Model Setup

The LLMs were accessed through the Hugging Face API and utilized in their quantized versions to reduce computational resource consumption and minimize memory usage. A maximum sequence length of 512 tokens was selected, as most comments are shorter than this threshold, ensuring better compatibility and efficient processing.

### 4.1.2 Dataset

The test dataset, which was previously separated from the training and validation sets, was used for this first experiment.

### 4.1.3 Process

A batch size of ten was used for batch inferencing. Padding, truncation, and a maximum token length of 512 were used during tokenization. To minimize computational overhead and disable gradient computations, inference was carried out under a `torch.no_grad()` context. For each batch, logs were gathered and kept for analysis.

### 4.1.4 Findings

In the first experiment, LLMs were evaluated using direct inferencing without any fine-tuning. The findings show a notable discrepancy between the actual and predicted label distributions, particularly for the minority class (toxic comments).

Most models performed quite near to actual counts for label 0 (non-toxic); however, a significant overestimation was noted for label 1 (toxic) in a number of models, including Falcon 7B, Meta-Llama-3 8B, and BERT multilingual uncased. This is summarized in Table 4.1.

**Table 4.1:** Actual vs. Predicted Comment Counts for Each Label

| Model | Actual (0) | Pred (0) | Actual (1) | Pred (1) |
|---|---|---|---|---|
| Mistral-7B-v0.3 | 4190 | 3910 | 774 | 1054 |
| DeepSeek-R1-7B | 4190 | 3632 | 774 | 1332 |
| Falcon-7B | 4190 | 2260 | 774 | 2704 |
| Llama-3-8B | 4190 | 1943 | 774 | 3021 |
| Gemma-2B | 4190 | 4797 | 774 | 167 |
| BERT Multilingual | 4190 | 1805 | 774 | 3159 |

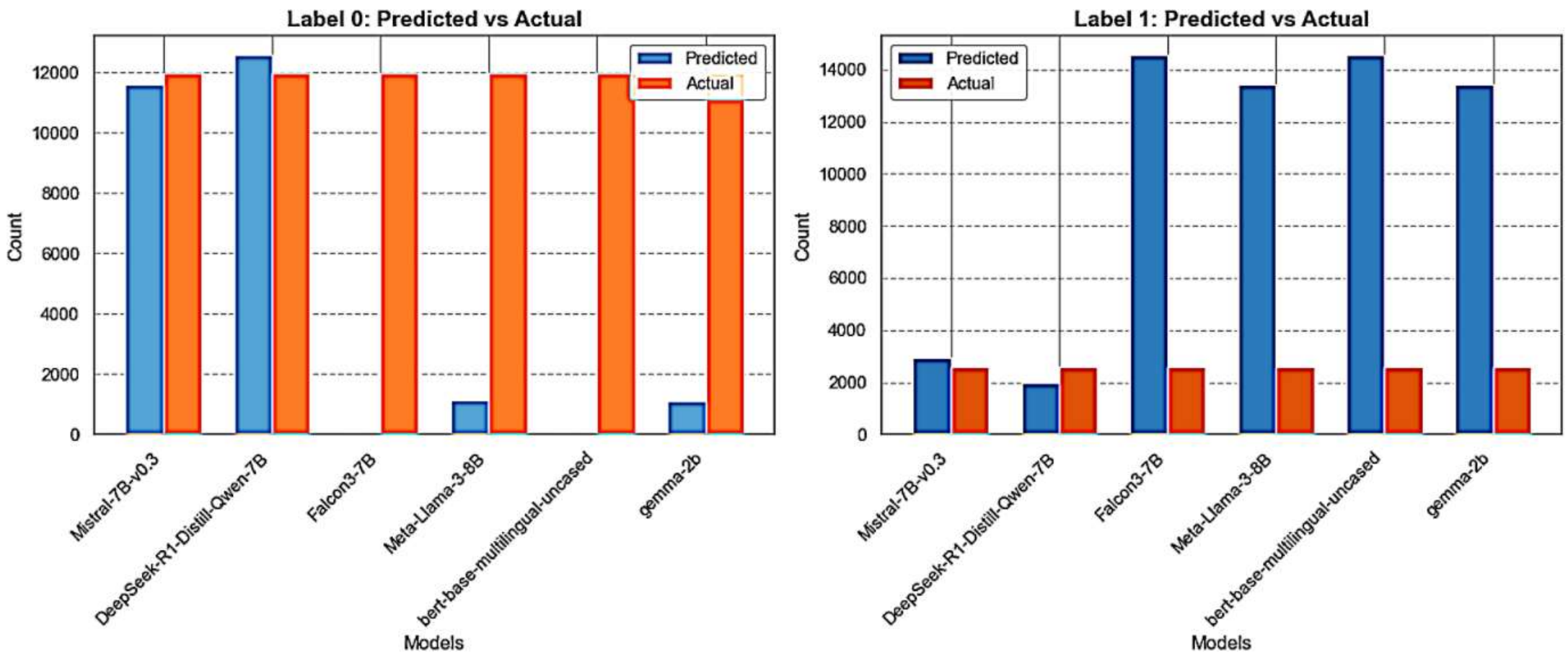

**Figure 4.1:** Predicted vs. actual comment counts for label 0 and label 1 across models.

Mistral-7B-v0.3 performed best overall in terms of evaluation metrics, with an accuracy of 0.7262 and a macro F1-score of 0.56. Then came DeepSeek-R1-Distill-Qwen-7B, which had an accuracy of 0.7402 and an F1-score of 0.51. Gemma-2B, Meta-Llama-3-8B, and BERT multilingual uncased were among the models that produced significantly lower F1-scores (0.21, 0.21, and 0.15, respectively), indicating that they were unable to correctly categorize harmful comments. These performance metrics are detailed in Table 4.2.

**Table 4.2:** Evaluation metrics (F1, Accuracy, Precision, Recall) across models for baseline LLM performance evaluation.

| Model | F1 | Accuracy | Precision | Recall |
|---|---|---|---|---|
| Mistral-7B-v0.3 | 0.56 | 0.7262 | 0.55 | 0.56 |
| DeepSeek-R1-7B | 0.51 | 0.7402 | 0.51 | 0.51 |
| Falcon-7B | 0.32 | 0.3178 | 0.51 | 0.51 |
| Llama-3-8B | 0.21 | 0.2217 | 0.48 | 0.49 |
| Gemma-2B | 0.21 | 0.2221 | 0.48 | 0.49 |
| BERT Multilingual | 0.15 | 0.1779 | 0.46 | 0.50 |

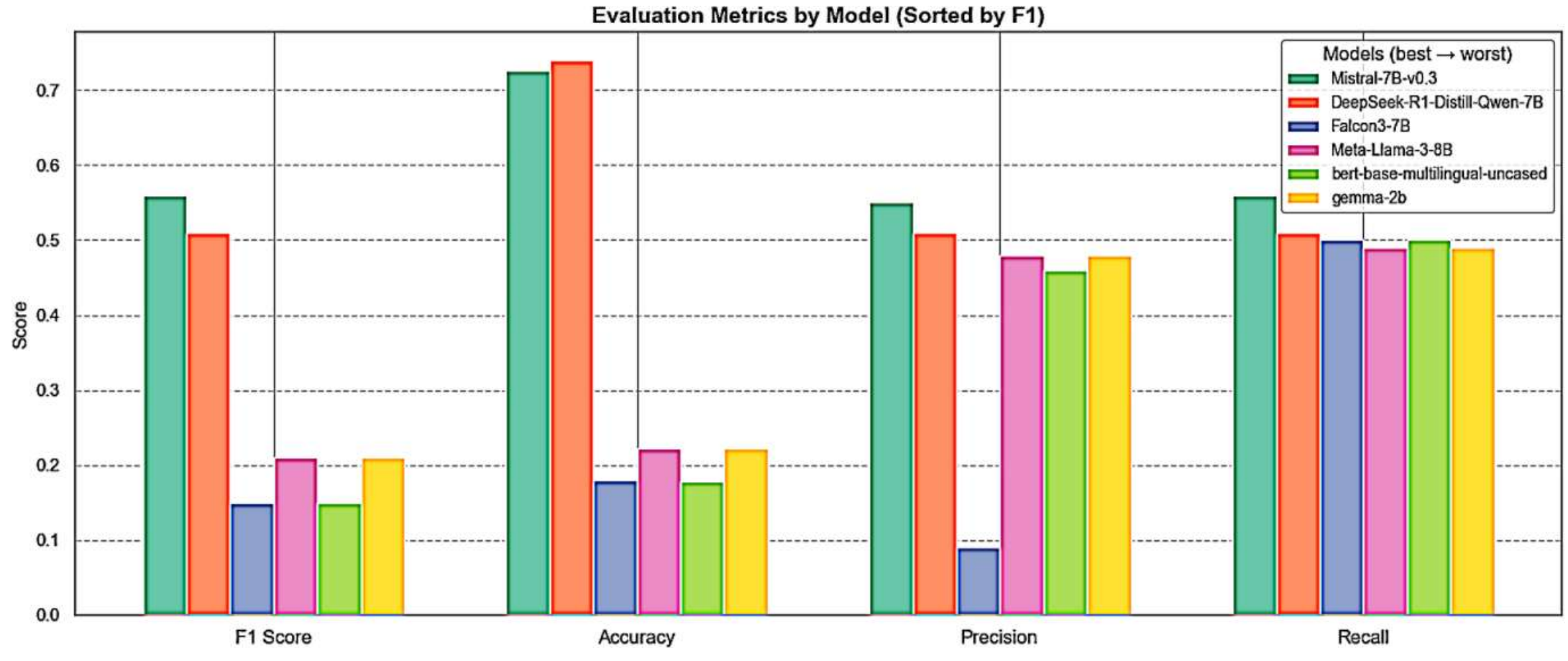

**Figure 4.2:** Evaluation metrics by model (sorted by F1 score).

While models tried to extract more harmful remarks, they frequently misclassified non-toxic samples,

raising false positives, according to precision and recall scores, which support these findings. Most models achieved better recall than precision.

These limitations are further highlighted by confusion matrix analysis. For example, BERT multilingual uncased, Falcon 7B, and Meta-Llama-3-8B all displayed extremely high false positive rates (FP = 9411, 11087, and 11962, respectively), which led to low precision even with decent recall. Mistral-7B-v0.3, on the other hand, preserved a more equitable trade-off between false positives (FP = 2177) and true positives (TP = 774). See Table 4.3.

**Table 4.3:** Confusion matrix metrics (TP, TN, FP, FN) across models for baseline LLM performance evaluation.

| **Model** | **TP** | **TN** | **FP** | **FN** |
|---|---|---|---|---|
| Mistral-7B-v0.3 | 774 | 3910 | 2177 | 1054 |
| DeepSeek-R1-7B | 388 | 3632 | 1588 | 1332 |
| Falcon-7B | 2063 | 2260 | 9411 | 2704 |
| Llama-3-8B | 2345 | 1943 | 11087 | 3021 |
| Gemma-2B | 2346 | 4797 | 11081 | 167 |
| BERT Multilingual | 2578 | 1805 | 11962 | 3159 |

All things considered, these results highlight the shortcomings of classifying hate speech in Roman Urdu using LLMs alone, without task-specific modification or fine-tuning. Additional strategies, like prompt tuning or parameter-efficient fine-tuning, are required to enhance minority class detection and overall performance due to the notable class imbalance and linguistic complexity.

## 4.2 LLM Performance Evaluation Post PEFT Fine-Tuning

This experiment fine-tunes each model with LoRA (freezing backbone weights and updating lightweight adapters) and then evaluates on the same held-out test split as the baseline. Training ran on an NVIDIA A100 with minimal hyperparameter tuning, and pre-/post-PEFT performance is compared under an identical evaluation protocol to quantify accuracy gains versus compute/parameter efficiency.

### 4.2.1 Model Setup

For each LLM, LoRA configurations were used, which enabled effective parameter adaptation without requiring updates to the entire model weights. The initial layers that had been pre-trained were still frozen.

### 4.2.2 Dataset

Results before and after fine-tuning were directly compared using the same test set (20% of the dataset) as in baseline LLM performance evaluation.

### 4.2.3 Process

NVIDIA A100 GPUs were used for fine-tuning. A total of three to seven hours were spent training each model over the course of two epochs. Only minor tweaks were required, however hyperparameters such batch size and learning rate were adjusted.
The architectural summary before and after LoRA adaptation is displayed in Table 4.4. All of the parameters are trainable at first, but only a small portion (about 7 million parameters) are updated after applying LoRA adapters. This significantly lowers the fine-tuning cost while maintaining the integrity of the entire architecture (decoder layers and backbone).

**Table 4.4:** Architecture summary before and after LoRA adaptation

| **Model** | **Total Params** | **Trainable Params Before LoRA** | **Trainable Params After LoRA** | **Decoder Layers** |
|---|---|---|---|---|
| Llama 3.2 8B | 8B | 8B | ∼7M | 32 |
| Gemma 2B | 2B | 2B | ∼6.6M | 18 |
| Mistral 7B | 7B | 7B | ∼7.2M | 32 |
| DeepSeek R1-Qwen 7B | 7B | 7B | ∼7.2M | 40 |
| Falcon 7B | 7B | 7B | ∼7.2M | 32 |
| BERT multilingual | 110M | 110M | ∼7M | 24 |

### 4.2.4 Findings

The LoRA-based parameter-efficient fine-tuning (PEFT) method was used to fine-tune LLMs before they were assessed in the second experiment. Model performance for both the majority (non-toxic) and minority (toxic) classes significantly improved as a result of this procedure. After fine-tuning,

Table 4.5 shows that the projected label distributions closely matched the actual distributions. Most models successfully represented the minority class without significantly misclassifying non-toxic comments, in contrast to the previous experiment.

**Table 4.5:** Actual vs. Predicted Comment Counts for Each Label (After Fine-Tuning)

| Model | Actual (0) | Pred (0) | Actual (1) | Pred (1) |
|---|---|---|---|---|
| Gemma-2B | 11973 | 11910 | 2582 | 2645 |
| BERT Multilingual | 11973 | 11933 | 2582 | 2622 |
| Meta-Llama-3-8B | 11973 | 12013 | 2582 | 2542 |
| Mistral-7B-v0.3 | 11973 | 11966 | 2582 | 2589 |
| DeepSeek-R1-7B | 11973 | 11194 | 2582 | 3361 |
| Falcon-7B | 11973 | 11194 | 2582 | 3361 |

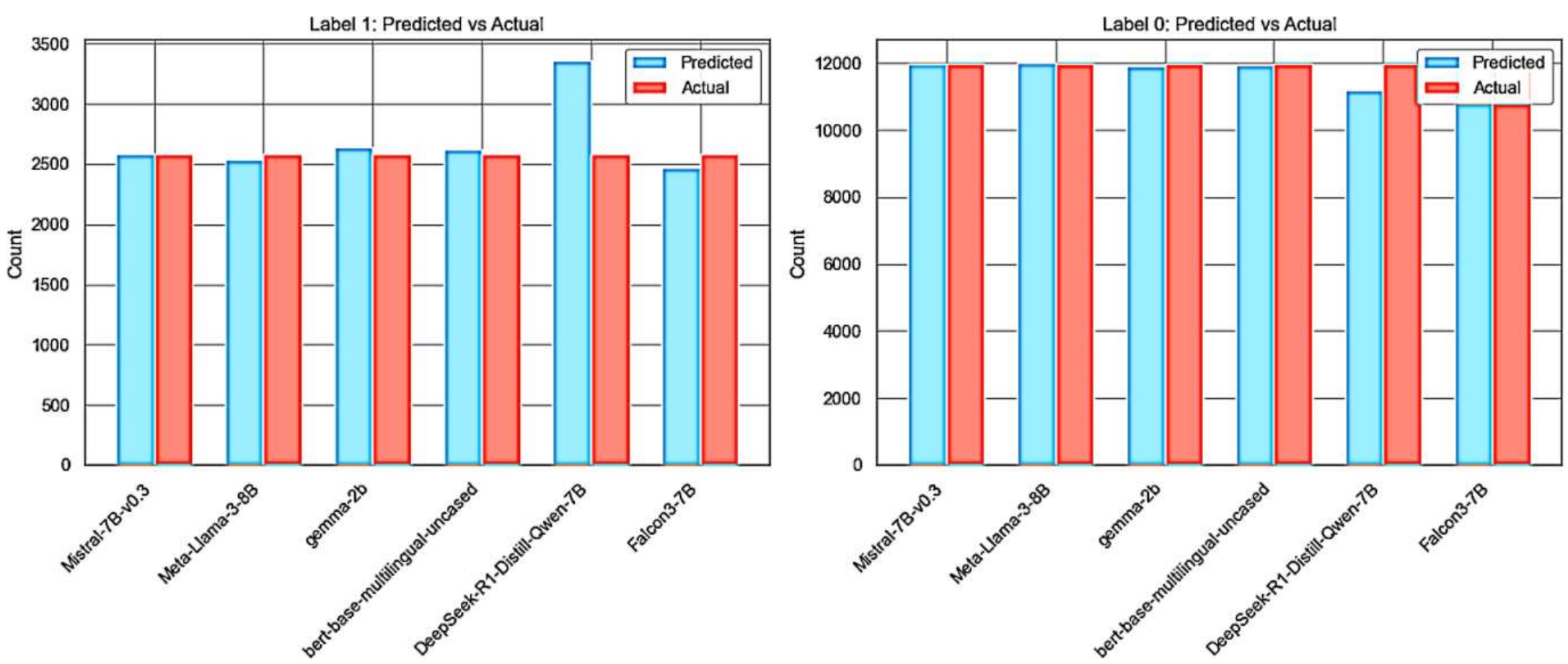

**Figure 4.3:** Predicted vs. actual comment counts per label after fine-tuning.

After being fine-tuned, all models showed a noticeable improvement in F1 results. Notably, F1 values above 0.93 were attained by Mistral-7B-v0.3 and Meta-Llama-3-8B, while other models also demonstrated notable improvements. The metrics are shown in full in Table 4.6.
In contrast to the previous experiment, most models were able to attain balanced precision and recall values after fine-tuning, as shown in Table 4.6. Additionally, the accuracy values show the general improvement in efficiently capturing both toxic and non-toxic classes.

**Table 4.6:** Evaluation metrics (F1, Accuracy, Precision, Recall) across models for LLM performance evaluation post PEFT fine-tuning.

| Model | F1 | Accuracy | Precision | Recall |
|---|---|---|---|---|
| Mistral-7B-v0.3 | 0.9387 | 0.9642 | 0.9383 | 0.9392 |
| Meta-Llama-3-8B | 0.9379 | 0.9640 | 0.9407 | 0.9353 |
| Falcon-7B | 0.9180 | 0.8354 | 0.7278 | 0.7773 |
| Gemma-2B | 0.9089 | 0.9463 | 0.9052 | 0.9129 |
| BERT Multilingual | 0.8203 | 0.8945 | 0.8184 | 0.8222 |
| DeepSeek-R1-7B | 0.7468 | 0.8354 | 0.7278 | 0.7773 |

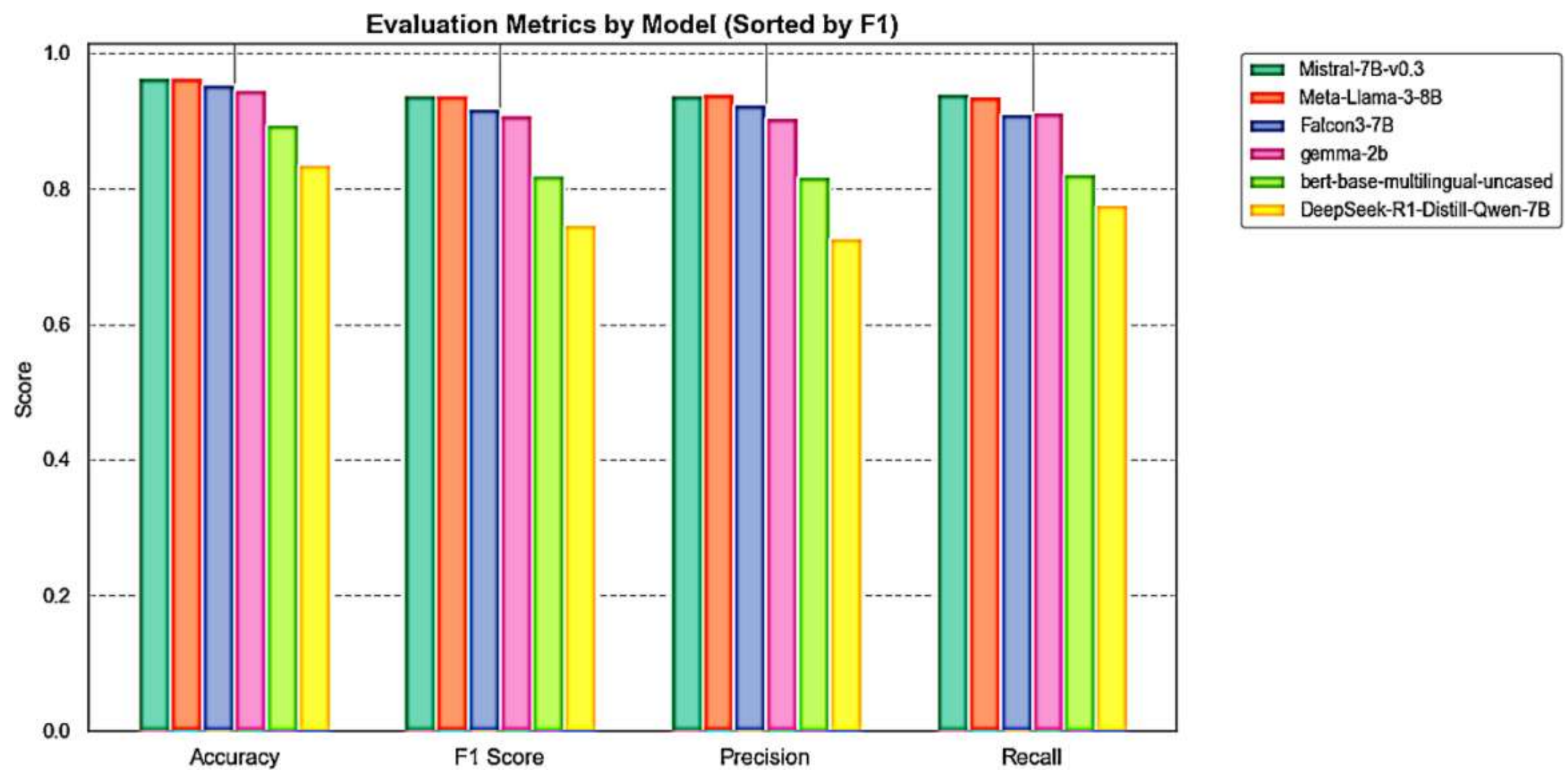

**Figure 4.4:** Evaluation metrics (Accuracy, F1 Score, Precision, Recall) for each model after fine-tuning.

Confusion matrix metrics are shown in Table 4.7, which further validates the enhanced performance. More robust generalization was indicated by the lowest false positive and false negative numbers for Mistral-7B-v0.3 and Meta-Llama-3-8B.

**Table 4.7:** Confusion matrix metrics (TN, FP, FN, TP) across models for LLM performance evaluation post PEFT fine-tuning.

| Model | TN | FP | FN | TP |
|---|---|---|---|---|
| Gemma-2B | 11551 | 422 | 359 | 2223 |
| BERT Multilingual | 11185 | 788 | 748 | 1834 |
| Meta-Llama-3-8B | 11731 | 242 | 282 | 2300 |
| Mistral-7B-v0.3 | 11709 | 264 | 257 | 2325 |
| DeepSeek-R1-7B | 10386 | 1587 | 808 | 1774 |
| Falcon-7B | 10386 | 1587 | 808 | 1774 |

Lastly, Table 4.8 summarizes the models' F1 scores before and after fine-tuning.

**Table 4.8:** F1 Scores Before and After Fine-Tuning (PEFT)

| Model | F1 Score Before | F1 Score After |
|---|---|---|
| Mistral-7B-v0.3 | 0.56 | 0.9387 |
| Meta-Llama-3-8B | 0.21 | 0.9379 |
| Falcon-7B | 0.32 | 0.9180 |
| Gemma-2B | 0.21 | 0.9089 |
| BERT Multilingual | 0.15 | 0.8203 |
| DeepSeek-R1-7B | 0.51 | 0.7468 |

Overall, our results show that the classification of hate speech in Roman Urdu is greatly improved by LoRA-based fine-tuning. The method produced balanced and higher F1 scores, decreased false positive rates, and significantly enhanced minority and majority class forecasts. For the robust and accurate deployment of LLMs in delicate applications like hate speech detection, fine-tuning is more important than direct inferencing.

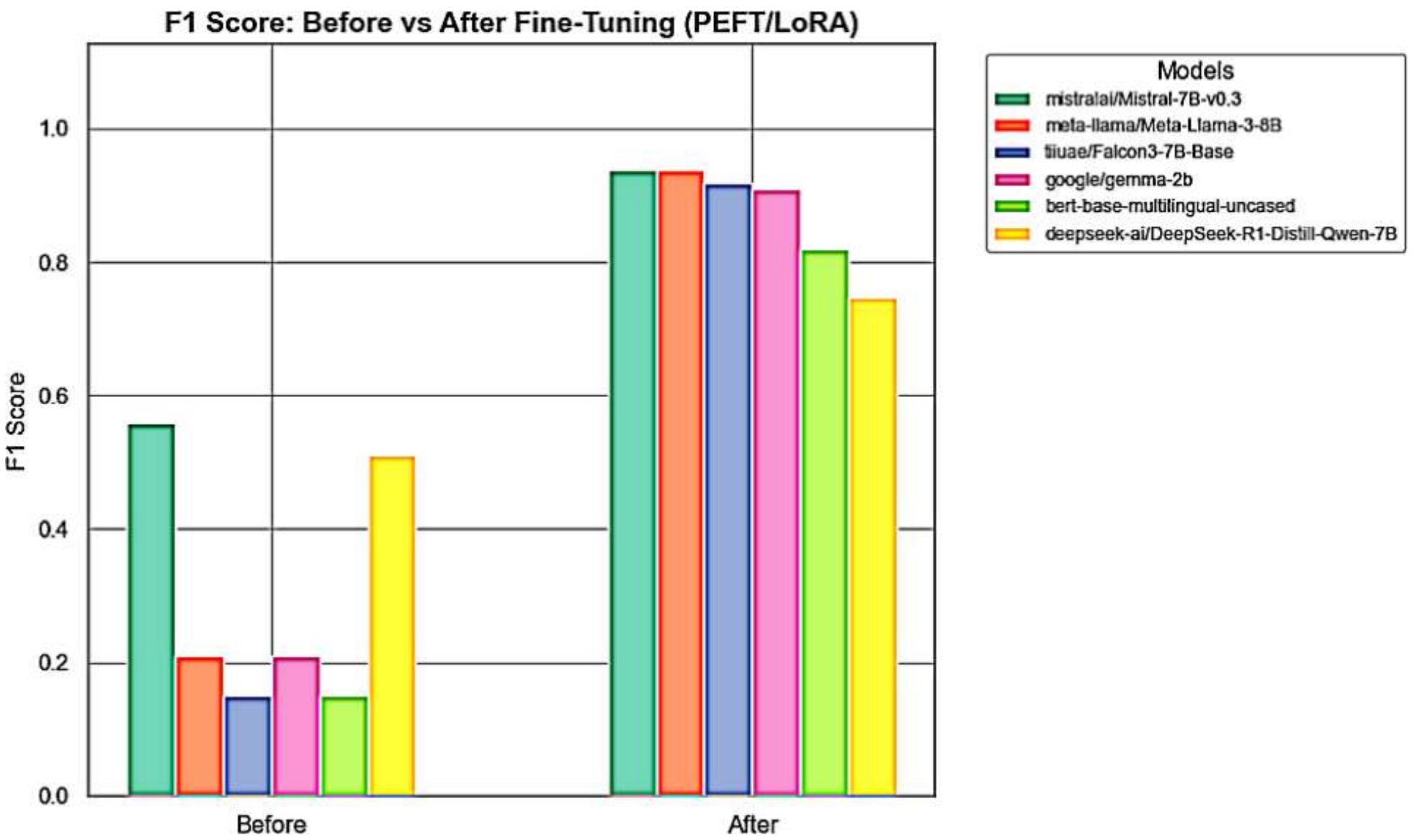

**Figure 4.5:** F1 scores before vs after fine-tuning using PEFT/LoRA (LLM performance evaluation post PEFT fine-tuning).

## 4.3 Performance Evaluation of BERT, PEFT with LoRA and Prompt Tuning

This experiment benchmarks BERT multilingual (uncased) under three strategies: direct inference, LoRA-based PEFT, and prompt tuning (prefix, cloze, and mixed) on disjoint train/dev/test splits. Prompt tuning is evaluated in few-shot regimes with $K \in \{32, 64, 128\}$, using two non overlapping rounds (two epochs each) per $K$. All runs use the same preprocessing and evaluation protocol, and the results are reported by precision, recall, precision, and F1 with tables detailing per $K$ and per prompt outcomes.

### 4.3.1 Approach

The BERT multilingual uncased model was tested in this experiment using several tuning methods to determine how well it classified hate speech in Roman Urdu. Three fundamental strategies were taken into consideration:

- Direct inferencing (without fine-tuning)
- Parameter-Efficient Fine-Tuning (PEFT) using LoRA
- Prompt tuning using manually crafted prompts

For prompt tuning, two types of manual prompt styles were implemented:

- Prefix prompts

- Cloze prompts

A mixed prompt approach, combining prefix and cloze styles, was also evaluated.

### 4.3.2 Data Setup

The dataset $D$ was divided into three disjoint subsets: training set $D_{\text{train}}$, development (validation) set $D_{\text{dev}}$, and test set $D_{\text{test}}$, such that:

$$D = D_{\text{train}} \cup D_{\text{dev}} \cup D_{\text{test}}, \quad D_{\text{train}} \cap D_{\text{dev}} = D_{\text{train}} \cap D_{\text{test}} = D_{\text{dev}} \cap D_{\text{test}} = \emptyset.$$

Three different configurations of training and development examples were considered with $K \in \{32, 64, 128\}$:

$$|D_{\text{train}}| = K, \quad |D_{\text{dev}}| = K, \quad |D_{\text{test}}| = |D| - 2K.$$

For each configuration, the examples were randomly sampled, and each configuration underwent two full rounds of training, validation, and testing over two epochs, i.e.,

$$\text{Rounds} = 2, \quad \text{Epochs per round} = 2,$$

ensuring no overlap of examples between rounds:

$$D^{(i)}_{\text{train}} \cap D^{(j)}_{\text{train}} = \emptyset, \quad \forall i \neq j.$$

### 4.3.3 Findings

Prompt tuning performance was initially examined in this experiment with varying numbers of training samples ($K$). As demonstrated in Tables 4.9, 4.10, and 4.11, metrics (precision, recall, F1 Score, and accuracy) were typically improved with greater values of $K$. This demonstrates how adding more varied instances improves the generalization and robustness of the model.
**B**elow is a summary of the evaluation metrics for each prompt type and $K$ value.

**Table 4.9:** Prefix Prompt Metrics by $K$

| K | Precision | Recall | F1 Score | Accuracy |
|---|---|---|---|---|
| 32 | 0.6311 | 0.5989 | 0.6095 | 0.7970 |
| 64 | 0.6222 | 0.6485 | 0.6312 | 0.7584 |
| 128 | 0.6747 | 0.6894 | 0.6813 | 0.8048 |

**Table 4.10:** Cloze Prompt Metrics by $K$

| K | Precision | Recall | F1 Score | Accuracy |
|---|---|---|---|---|
| 32 | 0.6097 | 0.6833 | 0.6532 | 0.6984 |
| 64 | 0.6342 | 0.7234 | 0.5886 | 0.6299 |
| 128 | 0.6256 | 0.7049 | 0.6167 | 0.6842 |

**Table 4.11:** Mixed Prompt Metrics by $K$

| K | Precision | Recall | F1 Score | Accuracy |
|---|---|---|---|---|
| 32 | 0.5992 | 0.6484 | 0.6008 | 0.6976 |
| 64 | 0.6405 | 0.6885 | 0.6531 | 0.7580 |
| 128 | 0.6774 | 0.7810 | 0.6852 | 0.7498 |

Figures 4.6, 4.7, and 4.8 show the detailed evaluation metrics for each prompt type (Prefix, Cloze, and Mixed) across different $K$ values (32, 64, 128). It is evident that higher $K$ values generally improve performance, with $K = 128$ achieving the best overall results in most cases.

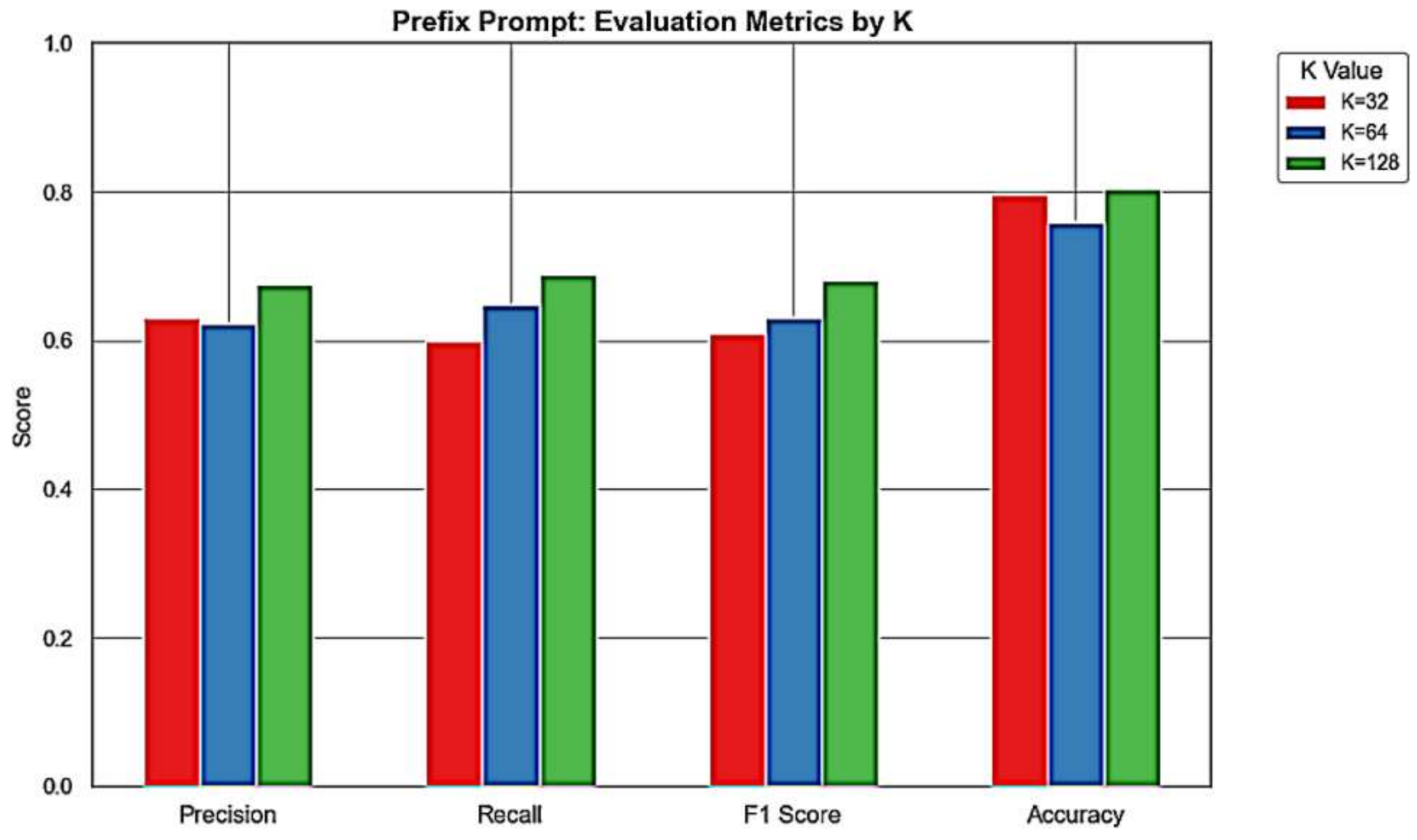

**Figure 4.6:** Prefix Prompt: Evaluation metrics across $K$ values.

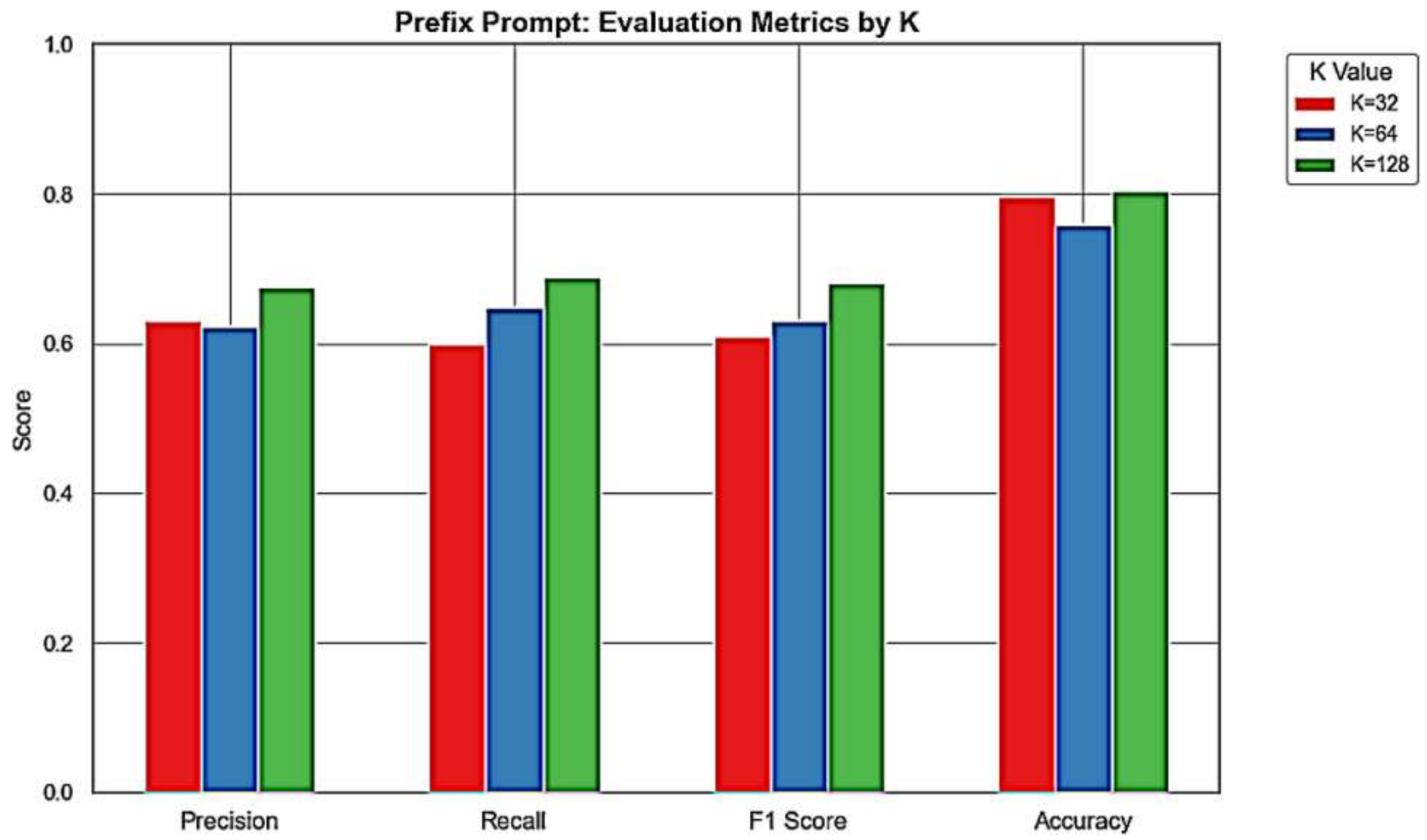

**Figure 4.7:** Cloze Prompt: Evaluation metrics across $K$ values.

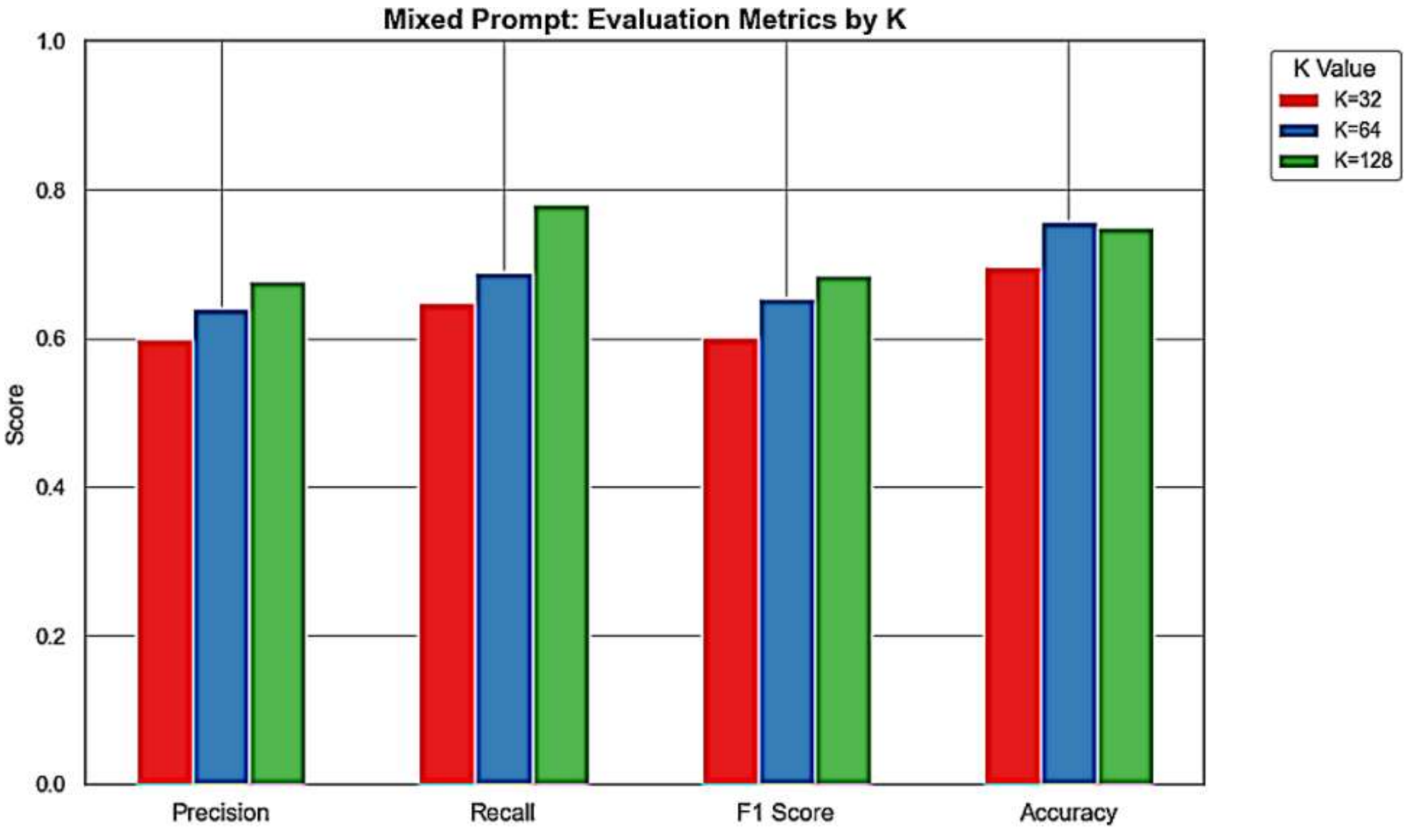

**Figure 4.8:** Mixed Prompt: Evaluation metrics across $K$ values.

The mixed prompts at $K = 128$ had the highest overall metric F1 Score of 0.6852, out of all configurations. Cloze prompts demonstrated comparatively poorer but still respectable performance at $K = 128$, but prefix prompts also performed competitively.

Table 4.12 summarizes the direct comparison of the best-performing metrics from each prompt type.

These findings demonstrate the relative advantages of each prompting technique in its optimal setting.

**Table 4.12:** Best Metrics for Different Prompt Types

| Prompt Type | Best F1 Score | Best Accuracy |
|---|---|---|
| Prefix | 0.6813 | 0.8048 |
| Cloze | 0.6167 | 0.6842 |
| Mixed | 0.6852 | 0.7580 |

Figure 4.9 highlights the best F1 score achieved by each prompt type, demonstrating that mixed and prefix prompts provided similar top-level performance, with mixed prompts slightly higher.

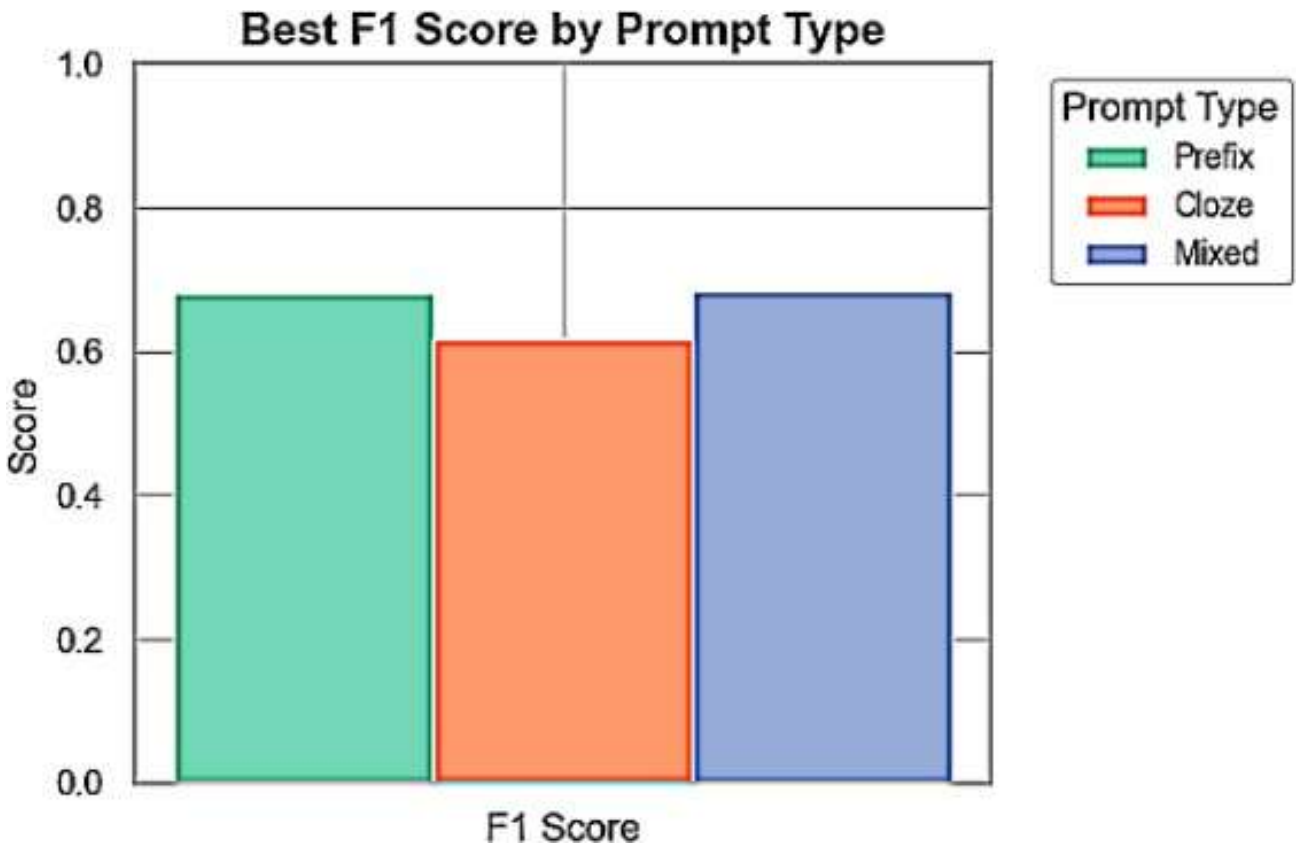

**Figure 4.9:** Best F1 score comparison among prompt types.

Finally, BERT was evaluated in three different settings: prompt tuning (taking into account optimal prompt configurations), PEFT with LoRA, and direct inference (without fine-tuning). The results in Table 4.13 indicate that PEFT LoRA had the highest accuracy (0.8945) and F1 score (0.82). Although prompt tuning approaches showed significant improvement over direct inference, they did not outperform PEFT LoRA, yet they provided lightweight substitute.

**Table 4.13:** F1 Score and Accuracy for Different BERT Techniques

| Technique | F1 Score | Accuracy |
|---|---|---|
| BERT (No Fine-Tuning) | 0.30 | 0.1779 |
| BERT + PEFT LoRA | 0.82 | 0.8945 |
| Mixed Prompt | 0.6852 | 0.7580 |
| Prefix Prompt | 0.6813 | 0.8048 |
| Cloze Prompt | 0.6167 | 0.6842 |

Figure 4.10 compares the F1 scores for each BERT-based technique including no fine-tuning, PEFT LoRA, and various prompt tuning styles.

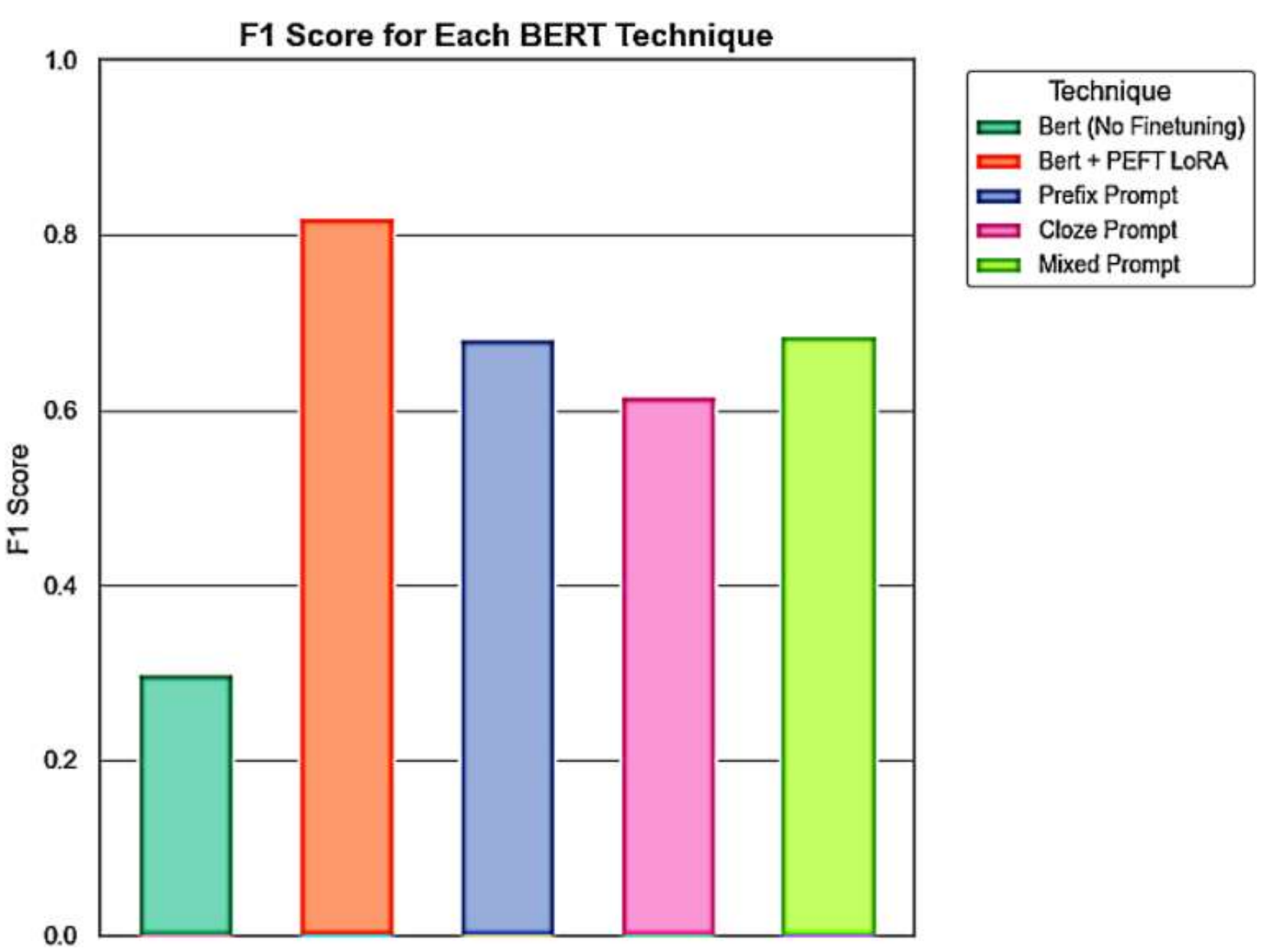

**Figure 4.10:** F1 score comparison for different BERT-based techniques.

These results highlight that prompt tuning provides a good compromise by enhancing performance without incurring full model fine-tuning expenses, especially when prefix and mixed styles are used. When optimal performance is needed, PEFT LoRA is still the best course of action.

## 4.4 Zero-Shot vs Few-Shot Prompt Engineering

Azure OpenAI `GPT-3.5` is benchmarked in this experiment for the classification of Roman Urdu toxicity under two prompting regimes: few shot (with labeled exemplars) and zero shot (instruction only). To enable a training-free, apples-to-apples comparison of the two prompting procedures, inference is performed in batches of 100 comments over the entire set. Metrics are calculated for each batch, and the final Accuracy, Precision, Recall, and F1 are provided as the mean across batches.

### 4.4.1 Approach

In this experiment, the OpenAI GPT-3.5 model, which could be accessed through Azure OpenAI services, was used to assess prompt engineering strategies. The goal was to evaluate the effectiveness of few-shot and zero-shot prompting techniques for Roman Urdu hate speech classification.
Two configurations were examined:

- **Zero-shot**: No explicit examples were provided; the model was only given task instructions.
- **Few-shot**: A small set of representative labeled examples were provided to guide predictions.

Data processing for both methods was done in batches of 100 comments per. To guarantee reliable performance measurements, the evaluation results from each batch were averaged to get the final evaluation metrics (Accuracy, Precision, Recall, and F1 Score).

### 4.4.2 Findings

The prompting techniques of zero-shot and few-shot showed good generalization capacities without the need for explicit fine-tuning. There were notable variations in performance, nevertheless.
All metrics showed that few-shot prompting performed better than zero-shot prompting. The few-shot method produced the greatest F1 score for one of the batches, which was 0.94939, while the zero-shot method produced an F1 score of 0.82503.
**Detailed final averaged evaluation metrics vs Maximum F1 Score are summarized in Table 4.14.**

**Table 4.14:** Final Averaged Evaluation Metrics vs Max. F1 Score for Prompt Engineering

| Prompt Type | Accuracy | Precision | Recall | F1 Score | Max F1 Score |
|---|---|---|---|---|---|
| Few-Shot | 0.9100 | 0.8278 | 0.8989 | 0.8565 | 0.9494 |
| Zero-Shot | 0.7229 | 0.6799 | 0.8129 | 0.6698 | 0.8250 |

Figure 4.11 illustrates the comparative evaluation metrics for Few-Shot and Zero-Shot prompt engineering. The figure clearly shows that Few-Shot prompts significantly outperform Zero-Shot prompts in all metrics: accuracy, F1 score, precision, and recall. This highlights the effectiveness of including a few carefully selected examples to guide the model in low-resource scenarios like Roman Urdu.

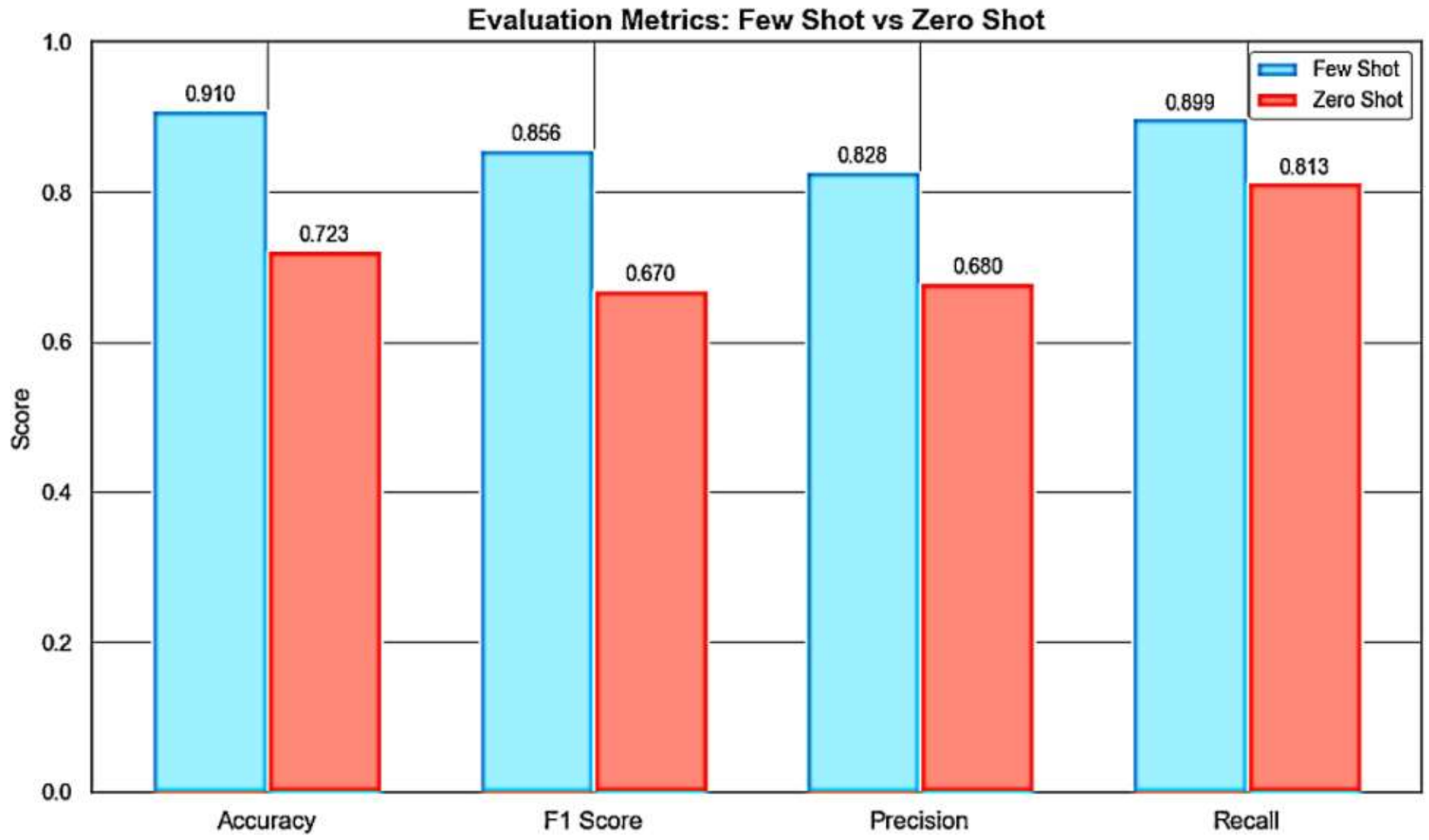

**Figure 4.11:** Evaluation metrics comparison between Few-Shot and Zero-Shot prompt engineering.

According to the findings, even a small amount of labeled instances helps the model comprehend task-specific patterns and subtleties, particularly in the intricate linguistic context of Roman Urdu. Its robustness is further confirmed by the graphical comparisons, which also show that few-shot prompting yields noticeably greater precision and F1 scores.

## 4.5 Consolidated Performance Analysis

To understand the overall performance landscape, results from all experiments and techniques were compared. This included:

- Direct inferencing with large language models (LLMs)
- Parameter-efficient fine-tuning (PEFT) using LoRA
- Prompt tuning (prefix, cloze, and mixed)
- Prompt engineering (zero-shot and few-shot)

Table 4.15 summarizes the average evaluation metrics across all techniques.

**Table 4.15:** Final Evaluation Metrics for All Techniques

| Technique | F1 Score | Accuracy | Precision | Recall |
|---|---|---|---|---|
| Mistral | 0.9387 | 0.9642 | 0.9383 | 0.9392 |
| LlaMA | 0.9379 | 0.9630 | 0.9407 | 0.9353 |
| Falcon-7B | 0.9200 | 0.9500 | 0.9300 | 0.9100 |
| Gemma | 0.9100 | 0.9463 | 0.9100 | 0.9100 |
| OpenAI Few-Shot | 0.8560 | 0.9100 | 0.8278 | 0.8988 |
| BERT (PEFT) | 0.8200 | 0.8945 | 0.8200 | 0.8200 |
| DeepSeek-Qwen | 0.7500 | 0.9354 | 0.7300 | 0.7800 |
| BERT Mixed Prompt | 0.6852 | 0.7498 | 0.6774 | 0.7810 |
| BERT Prefix Prompt | 0.6813 | 0.8048 | 0.6747 | 0.6894 |
| OpenAI Zero-Shot | 0.6698 | 0.7229 | 0.6799 | 0.8129 |
| BERT Cloze Prompt | 0.6167 | 0.6842 | 0.6256 | 0.7049 |

Figure 4.12 summarizes the overall F1 scores achieved by all models and techniques, ranked in descending order.

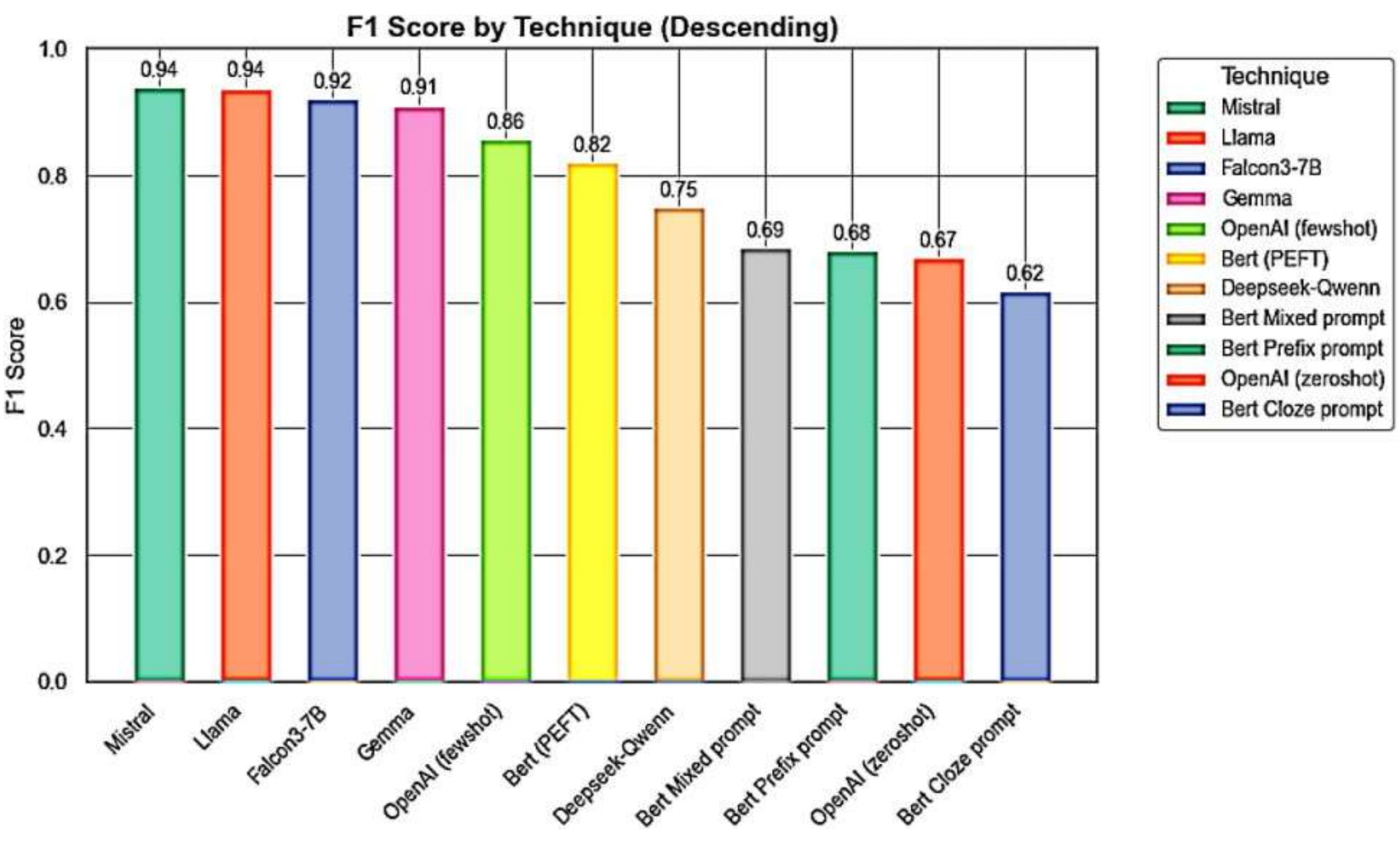

**Figure 4.12:** F1 Score by technique (descending) for all models and methods.

From this comprehensive analysis, it can be deduced that:

- The performance of large models such as LlaMA and Mistral, as well as parameter-efficient fine-tuning (PEFT) techniques, was exceptional, especially in terms of accuracy and F1 score.
- Using OpenAI Azure, few-shot rapid engineering techniques produced remarkably high peak F1 scores (up to 0.94939 in particular batches), sometimes almost surpassing PEFT, while average scores stayed marginally lower.
- Particularly at higher training example counts ($K = 128$), prefix and mixed prompts outperformed cloze prompts among prompt tuning techniques for BERT.
- Zero-shot prompting performed reasonably well in the absence of labeled data, although it was still less successful than PEFT and few-shot designs.

# Chapter 5

# Conclusion and Recommendations

*This chapter provides an overview of the key results from studies on the classification of hate speech in Roman Urdu using various techniques, such as prompt engineering, parameter-efficient fine-tuning (PEFT), and prompt tuning using large language models (LLMs). It also demonstrates the performance and adaptability of these techniques and offers suggestions for future lines of investigation.*

## 5.1 Conclusion

This thesis's thorough experimental investigation provided some crucial insights into how various strategies performed. In certain cases, prompt engineering can achieve higher F1 scores than parameter-efficient fine-tuning (PEFT) strategies. However, when generalized to larger datasets, its overall performance may become less consistent and unpredictable. Prompt engineering and prompt tuning can be a rapid and resource-efficient substitute for full model fine-tuning in situations when datasets are small or when there are few labeled examples available.

On the other hand, PEFT methods like LoRA typically produce more reliable and consistent results for bigger datasets or more complicated classification tasks, especially in low-resource languages. These techniques provide a better balance between computational efficiency and performance and better support robust generalization. All things considered, even if prompt-based techniques offer adaptability and flexibility, PEFT is still the best option in situations demanding great stability and accuracy, particularly when it comes to detecting hate speech in Roman Urdu or any other sequence classification tasks.

## 5.2 Recommendations

In light of the experimental results, the following suggestions are put out for further research. Investigate automated or semi-automated prompt crafting techniques to lessen reliance on human prompt design, which may improve prompt engineering's performance and scalability. Examine hybrid approaches that combine prompt-based techniques and PEFT (like LoRA) to maximize the benefits of both of these approaches for better results. To validate the generalization of the suggested methods in multilingual or cross-lingual contexts, extend the dataset, perform more tasks and consider other low-resource languages. These approaches seek to overcome the difficulties brought on by language diversity and resource limitations while enhancing the flexibility and efficacy of sequence classification models, especially hate speech detection.